\pdfoutput=1
\documentclass{article}

\usepackage[main, final]{neurips_2026}

\usepackage[utf8]{inputenc}
\usepackage[T1]{fontenc}
\usepackage{hyperref}
\usepackage{url}
\usepackage{booktabs}
\usepackage{amsfonts}
\usepackage{nicefrac}
\usepackage{microtype}
\usepackage{xcolor}
\usepackage{colortbl}
\definecolor{vmaecolor}{RGB}{215,232,250}   
\definecolor{dismocolor}{RGB}{225,243,220}  
\definecolor{ourscolor}{RGB}{255,236,200}   
\usepackage{amsmath}
\usepackage{amssymb}
\usepackage{graphicx}
\usepackage{adjustbox}
\usepackage{enumitem}
\usepackage{wrapfig}
\usepackage{placeins}
\usepackage{titletoc}

\title{TT-VidT: Decoupling the Temporal Axis for Efficient Motion-Centric Video Pretraining}

\newcommand{\aff}{\fontsize{10pt}{12pt}\selectfont}
\newcommand{\email}{\fontsize{9pt}{10pt}\selectfont}
\author{
  \textbf{Shih-Ying Yeh}$^{\spadesuit\heartsuit\dagger}$ \quad
  \textbf{Daniel Z.~Kaplan}$^{\diamondsuit}$ \quad
  \textbf{Xuehai Wang}$^{\clubsuit\triangle}$ \\
  \textbf{Fu-En Yang}$^{\bigstar}$ \quad
  \textbf{Min-Hung Chen}$^{\bigstar}$ \quad
  \textbf{Shang-Hong Lai}$^{\spadesuit}$
  \\[1.2em]
  \aff $^\spadesuit$National Tsing Hua University \enspace $^\heartsuit$Comfy Org Research \enspace $^\diamondsuit$realiz.ai \\
  \aff $^\clubsuit$Karolinska Institutet \enspace $^\triangle$Stockholm University \enspace $^\bigstar$NVIDIA \\[0.8em]
  $^\dagger$Corresponding author: \email{\texttt{kohaku@kblueleaf.net}} \\[0.3em]
  Project page: \url{https://kohakublueleaf.github.io/TTVidT/}
}

\begin{document}

\maketitle

\begin{abstract}
Comparisons in video self-supervised learning often evaluate complete training recipes rather than isolating the method itself: architecture, objective, data exposure, schedule, scale, and decoder capacity can all vary at once. This makes it hard to identify which choices yield motion-prioritized representations, whose gains concentrate on frame-to-frame change while retaining useful appearance. We address this with a matched $4 \times 6 = 24$ architecture-objective study at roughly 170M$\sim$190M encoder scale on $\sim$1.7M OpenVid and Moments-in-Time v2 clips for 8 epochs, and propose TT-VidT. TT-VidT combines a DINOv3-initialized ViT-B/16 per-frame spatial path with a compact Temporal Transfer Layer, trained by Diff Compression to reconstruct target frames from a first-frame appearance anchor and frame-specific motion tokens. The sweep shows that TT3D with Diff Compression, not either component alone, enters the strongest motion-sensitive regime, and decoder ablations favor a compact video-pretrained decoder. In final comparison, TT-VidT leads Jester, Something-Something\,V2, ARID, and Diving48 fine-tuning simultaneously, improving over the strongest non-TT row by 54\%$\sim$121\%, while using 48\% fewer encoder FLOPs than DisMo and 55\% fewer than VideoMAE or V-JEPA~2. HMDB51, IARD, and EPIC-Kitchens bound the claim.
%
\end{abstract}



\section{Introduction}
\label{sec:intro}

To be meaningfully different from frame understanding, video understanding must use information that no single frame contains, and we call the information recoverable from a single frame \emph{appearance}. In practice, however, many action-recognition benchmarks contain strong appearance cues: objects, scenes, actors, clothing, and camera context can often predict the label \citep{liu2021noframe, li2018resound, kowal2022staticdynamic, fioresi2025albar}. A video self-supervised learning (SSL) method can therefore obtain competitive recognition accuracy while relying heavily on appearance. This raises a more specific question: under a controlled comparison, which architecture-objective combinations produce representations whose gains concentrate on motion-sensitive tasks?

Existing video SSL methods approach this problem through three families: masked reconstruction (e.g., VideoMAE \citep{tong2022videomae}), latent prediction (e.g., V-JEPA, V-JEPA~2 \citep{bardes2024vjepa,assran2025vjepa2}), and motion-aware training (e.g., DisMo \citep{resslerantal2025dismo}). These approaches are effective, but prior comparisons often entangle architecture, objective, data exposure, schedule, and parameter scale. As a result, it is difficult to attribute motion-oriented behavior to a specific architecture-objective choice rather than to incidental recipe differences.

These observations suggest that temporal representation learning should be evaluated not only by average downstream accuracy, but also by whether a model improves in regimes where static appearance is insufficient. This distinction is difficult to isolate with conventional video SSL encoders, since spatial appearance and temporal evidence are typically mixed throughout the network \citep{arnab2021vivit, bertasius2021timesformer}. A model may therefore perform well on a video benchmark while relying heavily on identity, scene, or object cues that are visible in a single frame \citep{li2018resound, kowal2022staticdynamic}. We use the term \emph{motion-prioritized} to describe the empirical pattern in which a video representation improves most clearly on tasks whose labels depend on frame-to-frame change, while remaining competitive on appearance-dominated tasks.

\begin{figure}[t]
  \centering
  \includegraphics[width=\linewidth]{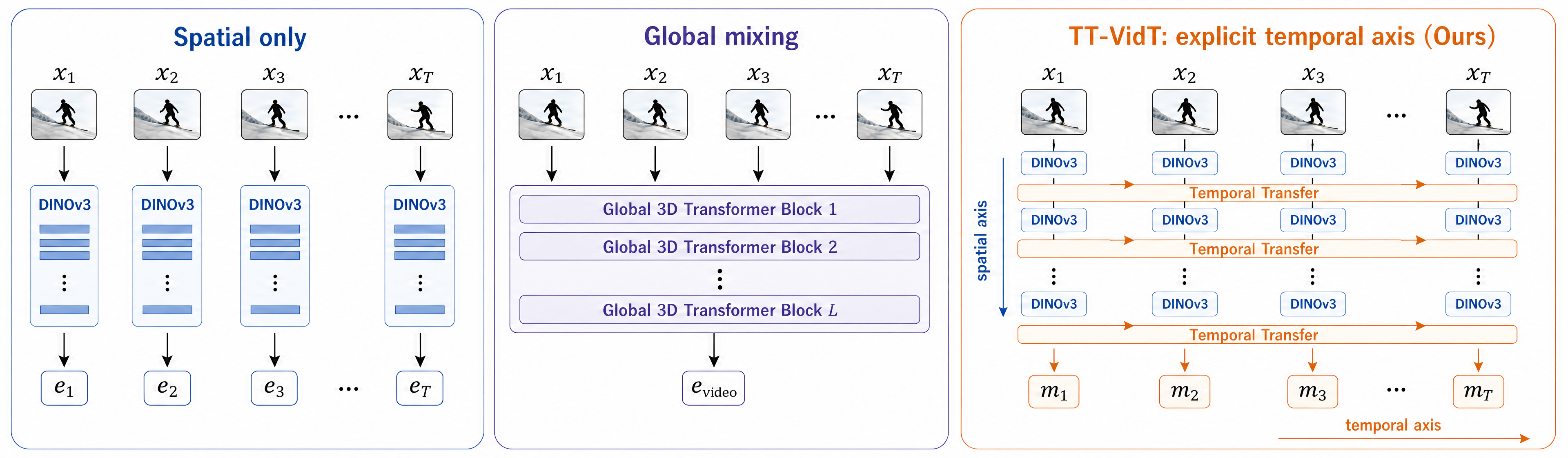}
  \caption{Overview of TT-VidT's decoupled pretraining design. A wide first-frame spatial feature supplies the appearance anchor, while a compact temporal transfer path emits frame-specific motion tokens for Diff Compression.}
  \label{fig:teaser}
\end{figure}

TT-VidT is designed to make this pattern measurable under a controlled recipe. The model keeps a wide per-frame spatial stream for appearance information, while routing temporal information through a compact transfer path (Figure~\ref{fig:teaser}). During pretraining, Diff Compression reconstructs target frames from a first-frame spatial anchor and a small set of frame-specific motion tokens. This formulation limits the capacity of the frame-specific path and encourages it to carry information that cannot be recovered from the anchor alone. In the temporal transfer layers, motion tokens are conditioned on compressed spatial context, while the high-capacity spatial anchor is reserved for reconstruction. This design does not remove appearance information from the model; rather, it makes the frame-specific bottleneck narrow enough that improvements on motion-sensitive tasks can be interpreted as evidence for useful temporal transfer under the matched recipe.

Our protocol pretrains all entries at a roughly 170M$\sim$190M encoder scale on a $\sim$1.7M clip mixture from OpenVid \citep{DBLP:conf/iclr/NanXZFYCL0T25} and Moments-in-Time v2 \citep{DBLP:journals/pami/MonfortVOAZRBYB20} for 8 epochs under a matched recipe, then evaluates $4 \times 6 = 24$ architecture-objective combinations against three canonical baselines. TT-VidT's profile concentrates on motion-heavy evaluations: $37.47$ on ARID, $73.25$ on Jester, $25.92$ on Something-Something\,V2, and $18.63$ on Diving48 fine-tuning. The profile reaches the representation itself: when the input motion is flipped or time-reversed, TT-VidT abandons its original answer on all but 1\% of clips while every baseline keeps it on 9$\sim$43\%, and the lead survives end-to-end finetuning, seeds, and a near-doubled budget (\S\ref{sec:exp-analysis}). HMDB51, IARD, and EPIC-Kitchens bound the claim where appearance favors broader baselines.

\paragraph{Contributions.}
\textbf{(1)} We introduce a controlled video SSL protocol that compares $24$ architecture-objective combinations under a shared recipe, supplemented by a single-frame appearance diagnostic to contextualize boundary cases.
\textbf{(2)} We propose TT-VidT, an instance combining TT3D and Diff Compression. TT3D adds a compact Temporal Transfer Layer on top of a DINOv3-initialized ViT-B/16 spatial path, jointly trained from these initial weights under the matched recipe; Diff Compression reconstructs target frames from a first-frame spatial anchor and a small set of motion tokens.
\textbf{(3)} We show that TT-VidT has an empirically motion-prioritized profile: it is the only method in our comparison group to lead Jester, Something-Something\,V2, ARID, and Diving48 fine-tuning simultaneously, with the gains confirmed per clip by a motion-inversion probe, while HMDB51, IARD, and EPIC-Kitchens expose the boundary of the claim.


\section{Related Work}
\label{sec:related-work}

\paragraph{Augmentation-based motion-appearance disentanglement.}
DisMo is the closest comparison both in scale and conceptual lineage. Our matched recipe trains all entries at a roughly 170M$\sim$190M encoder scale on $\sim$1.7M video clips for 8 epochs ($\sim$13.6M samples seen), close to DisMo's reported 172M / $\sim$17M setting \citep{resslerantal2025dismo}. DisMo learns a motion extractor and a motion-conditioned generator, using appearance augmentation to reduce identity leakage while preserving motion conditioning.
Native DisMo uses a DINOv2-B frame encoder and a 3D ViT-B sequence embedder that produces motion tokens \citep{oquab2023dinov2, resslerantal2025dismo}. In our matched DisMo-style row, we use the same DINOv3-initialized 2D path plus 3D transformer-block setup as TT-VidT where applicable, but with DisMo's motion-control objective rather than Diff Compression \citep{simeoni2025dinov3}. The resulting comparison isolates a narrow delta: DisMo tests augmentation-based motion control, while TT-VidT combines a compact temporal path with Diff Compression, with capacity pressure motivated by information-bottleneck views \citep{tishby2000informationbottleneck, alemi2017deepvib, achille2018informationdropout}.

\paragraph{Masked and autoregressive video pretraining at comparable scale.}
VideoMAE and ARVideo are near peers by ViT-B-scale architecture, but their data exposure is much larger than the matched recipe: VideoMAE accounts for 343M total parameters and about 410M seen clips on Kinetics-710 in our comparison accounting, while ARVideo accounts for 304M total parameters and roughly 406M seen clips under its Something-Something\,V2 schedule \citep{tong2022videomae, openmmlab2024kinetics710, ren2024arvideo}. Unlike DisMo's augmentation-based motion control, these methods train full video encoders with masked reconstruction or autoregressive token prediction, so architecture and objective comparisons remain informative although the exposure gap prevents a one-to-one controlled comparison.
Motion-aware MAE variants sharpen this neighborhood: AdaMAE, MAM2, MotionMAE, MME, SMILE, TrackMAE, and No More Shortcuts alter mask selection, split appearance and motion decoders, reconstruct temporal differences or trajectories, inject synthetic motion, or remove local appearance shortcuts \citep{bandara2022adamae, song2022mam2, yang2022motionmae, sun2023mme, thoker2025smile, vandeghen2026trackmae, dave2024nomoreshortcuts}. They validate the need for temporal targets, while DisMo and TT-VidT help contextualize appearance leakage in evaluation.

\paragraph{Large-scale latent-prediction and masked-video systems.}
V-JEPA, V-JEPA~2, and VideoMAE v2 define the latent-prediction and masked-video scale context, with V-JEPA~2 moving into VM22M-scale data and ViT-L to ViT-g encoders \citep{bardes2024vjepa, assran2025vjepa2, wang2023videomaev2}. Toto, VideoMAP, NExT-Vid, and SALT extend the frontier through autoregressive video pretraining, Mamba-Transformer hybrids, next-frame objectives, or static-teacher latent training \citep{rajasegaran2025toto, liu2025videomap, li2025nextvid, li2025salt}.
These systems set useful upper-bound context and baseline families, but their native recipes differ in model size, data volume, schedules, and often teacher compute. Even at our matched-recipe scale, close to DisMo's, TT-VidT is framed as a small-recipe design study; full-scale V-JEPA~2, VideoMAE v2, Toto, VideoMAP, NExT-Vid, and SALT operate in different regimes.

\paragraph{Image-pretrained substrates with temporal modules.}
A second lineage reuses strong image-pretrained features and adds video-specific temporal computation. AdViSe trains a lightweight R3D temporal module on top of an image foundation model with a playback-rate perception objective, while FRAME, SALT, and MVD reuse DINO, CLIP, image, or video teacher features for temporally consistent representation learning or distillation \citep{wu2025advise, tv2025frame, li2025salt, wang2023mvd}.
ST-Adapter, AIM, DiST, and ZeroI2V adapt image transformers for supervised image-to-video transfer through temporal adapters, joint adapters, distillation, or efficient transfer modules \citep{pan2022stadapter, yang2023aim, qing2023dist, li2024zeroi2v}. We cite these methods as architectural precedent for a DINOv3-initialized 2D path plus a narrow temporal pathway, with all parameters trained jointly, rather than as SSL pretraining baselines under the matched recipe.

\paragraph{Benchmarks, appearance bias, and diagnostic evaluation.}
Action-recognition benchmarks differ in motion demand, so the evaluation is written as a diagnostic ladder rather than a single leaderboard. HMDB51 is classic but biased toward scene, background, and subject appearance \citep{kuehne2011hmdb, liu2021noframe, fioresi2025albar}. Jester, Something-Something\,V2, and Diving48 stress hand motion, temporal order, human-object interaction, or fine-grained dynamics, yet static-dynamic analyses show that appearance cues can remain predictive \citep{materzynska2019jester, goyal2017something, li2018resound, kowal2022staticdynamic}.
ARID reduces appearance reliability through low-light capture, IARD tests identity-controlled invariance, and EPIC-Kitchens-100 mixes motion-heavy verbs with noun and scene signals \citep{xu2020arid, tacchetti2017iard, isik2017fast, damen2022epickitchens100}. SEVERE, SEVERE++, and VSSL benchmarking argue that single-score summaries hide domain, scale, granularity, and probe-capacity variation \citep{thoker2022severe, thoker2025severeplusplus, kumar2024benchmarkingvssl}. Each entry in our sweep is trained under the shared video pretraining recipe; image-path entries (TT3D, TT1D, and DisMo) initialize the spatial encoder from DINOv3 ViT-B/16, while VideoMAE-style 3D ViT entries are trained from scratch. All entries are paired with a single-frame DINOv3 appearance control \citep{simeoni2025dinov3}; the probe ladder combines kNN, linear and MLP probes over mean-pooled or learned-weight features, attentive aggregation, and DisMo-style identity checks \citep{caron2021dino, chen2020simclr, he2022mae, bardes2024vjepa, resslerantal2025dismo}.

\section{Method}
\label{sec:method}

We present TT-VidT, a video self-supervised method built from two parts. The encoder pairs a DINOv3-initialized 2D spatial path with a compact \emph{Temporal Transfer} pathway that produces per-frame motion embeddings; we instantiate it in two forms, \emph{TT1D} (Section~\ref{sec:method-tt}) and the joint space-time variant \emph{TT3D} (Section~\ref{sec:method-tt3d}). The pretraining objective is \emph{Diff Compression} (Section~\ref{sec:method-diffcomp}): the decoder reconstructs each later frame from a single first-frame appearance anchor paired with a per-target-frame motion token, so the encoder must place into each motion token whatever frame-specific information the decoder needs beyond the anchor.

We use 1-indexed frames throughout: $X=(x_1,\dots,x_T)$ with $T=8$, $z_t=E(x_t)\in\mathbb{R}^{N\times d}$ from a DINOv3 ViT-B/16 encoder $E$ \citep{simeoni2025dinov3} with $N=256, d=768$, $m_t\in\mathbb{R}^{K\times d}$ with $K=8$ motion tokens per frame, decoder $D$, and per-frame loss $\ell$. All parameters are trained jointly under the matched recipe.

\subsection{Temporal Transfer}
\label{sec:method-tt}

The temporal channel of TT-VidT is built around a small set of motion tokens that summarize per-frame spatial context and exchange information across frames. The spatial path is the standard DINOv3-initialized ViT applied independently per frame, with no cross-frame mixing.

We attach $K=8$ learnable motion-token embeddings to each frame. The Temporal Transfer Layer $\mathcal{T}$ has $L_{\mathcal{T}}=12$ layers with hidden dimension $d=768$. In each layer, the motion tokens of frame $t$ are concatenated with that frame's spatial tokens and processed by the self-attention of the ViT layer, so each motion group summarizes its own frame. The motion tokens then attend to each other under a block-causal mask: the $T\cdot K$ motion tokens form a single sequence in which frame $t$ may attend to frames $1,\dots,t$. After the final layer, the $T$ groups $\{m_1,\dots,m_T\}$ are read out for the pretraining objective (Section~\ref{sec:method-diffcomp}).

The cross-frame attention has length $T\cdot K = 64$, much smaller than the $T\cdot N = 2048$ that running attention over full spatial features would require. This keeps the matched comparison setting feasible, where data, schedule, and parameter scale are held constant across the sweep so that the properties of objective and encoder can be isolated. We refer to this configuration as \emph{TT1D}, since the cross-frame mixing runs purely along the time dimension over motion tokens.

\subsection{TT3D: Joint Spatial-Temporal Mixing}
\label{sec:method-tt3d}

TT3D is a variant of Temporal Transfer in which the cross-frame step sees more than the motion tokens. Instead of letting only the motion tokens exchange information across frames, we admit a coarse view of the spatial features directly into the cross-frame attention.

In each layer of $\mathcal{T}$, spatial tokens are reduced from $N=256$ to $N'=16$ per frame by a $4{\times}$ per-axis downsample, then concatenated with the $K=8$ motion tokens of that frame. A single block-causal 3D self-attention runs over the joint per-frame sequence of $K+N'=24$ tokens, with frame $t$ attending to frames $1,\dots,t$. Motion tokens and downsampled spatial tokens therefore share one temporal attention, so the motion tokens read spatial context from every earlier frame. The motion outputs are read out as in TT1D and consumed by the pretraining objective (Section~\ref{sec:method-diffcomp}).

The joint sequence has length $T\cdot(K+N') = 192$, still well below $T\cdot N = 2048$. The decoder continues to cross-attend to the full $z_1$; the downsample is local to $\mathcal{T}$ and does not propagate to the appearance pathway used by Diff Compression.

\begin{figure*}[t]
  \centering
  \includegraphics[width=\textwidth]{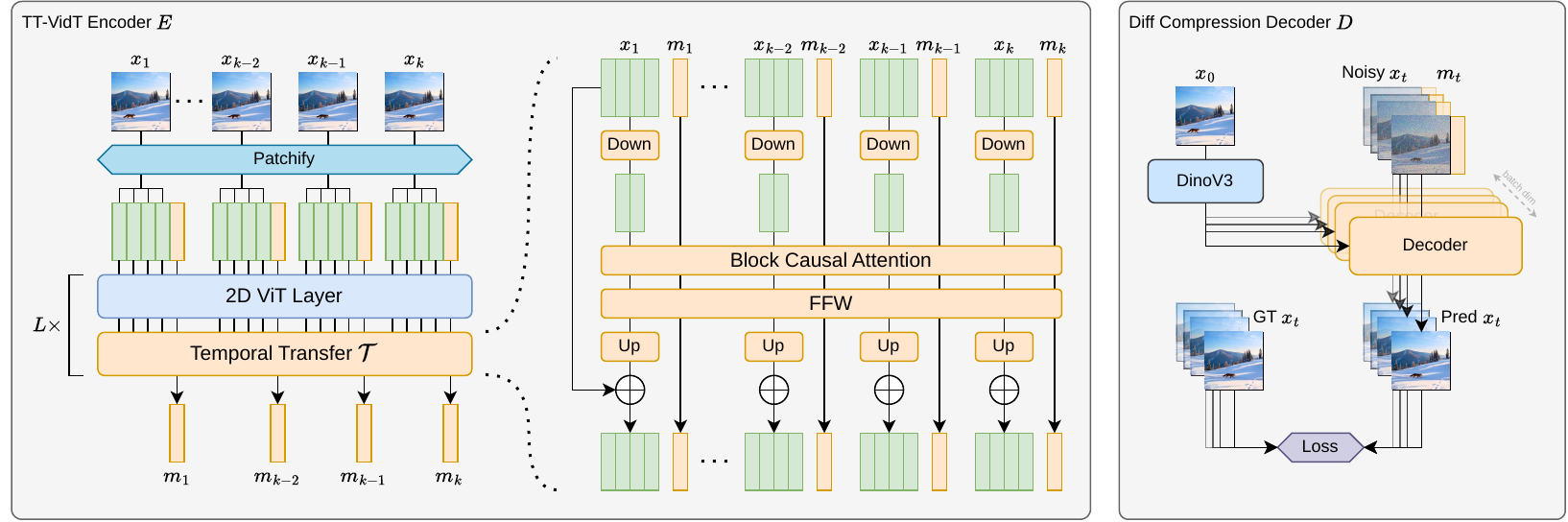}
  \caption{TT-VidT pretraining overview. The encoder interleaves DINOv3 2D ViT processing with Temporal Transfer. The Diff Compression decoder performs diffusion-style denoising conditioned on the first-frame anchor and motion tokens, with loss computed against the target frame.}
  \label{fig:tt_arch}
\end{figure*}

\subsection{Diff Compression}
\label{sec:method-diffcomp}

Diff Compression takes the first frame's spatial features as appearance anchor and the $t$-th frame's motion embedding as the carrier of frame-specific information:
\begin{equation}
\hat{x}_t = D(z_1,\,m_t), \qquad t = 2, \dots, T.
\label{eq:diffcomp}
\end{equation}
The per-frame loss is $\mathcal{L}_t=\ell(\hat{x}_t,x_t)$ with $\ell$ a diffusion or flow-matching reconstruction loss \citep{peebles2023dit, DBLP:conf/iclr/LipmanCBNL23} by default and latent regression as an ablation. The pretraining loss averages over non-anchor frames, $\mathcal{L}(X) = \frac{1}{T-1}\sum_{t=2}^{T}\mathcal{L}_t$, and the encoder $E$, Temporal Transfer Layer $\mathcal{T}$, decoder $D$, and motion-token embeddings are all trained jointly; recipe details are deferred to Section~\ref{sec:exp-setup}.

The factorization inspired by VTok \citep{wang2026vtokunifiedvideotokenizer}: a single key frame $z_1$ is paired with per-target-frame motion tokens $m_t$ for $t=2,\dots,T$, rather than a frame-by-frame autoregressive chain. We differ from VTok in how each $m_t$ is produced. VTok computes it explicitly, by a feature subtraction between the target frame and the key frame followed by a projection. In our setup $m_t$ is the direct output of the Temporal Transfer pathway, learned end-to-end under the reconstruction objective; the architecture contains no built-in feature diff, so the encoder is free to place into $m_t$ whatever information lets $D(z_1, m_t)$ reproduce $x_t$. The decoder cross-attends to the full $z_1\in\mathbb{R}^{N\times d}$, not the $N'$-downsampled features used inside $\mathcal{T}$ (Section~\ref{sec:method-tt3d}); the appearance pathway stays wide while the motion pathway remains compact. We hold the decoder at a compact S configuration; ablations in Section~\ref{sec:exp-decoder} show that scaling decoder capacity degrades motion-focused performance.

On the same encoder, MAE-Diffusion underperforms Diff Compression in our sweep, identifying the first-frame appearance anchor as the operative difference (Table~\ref{tab:sweep}). The pairing also exhibits a coupling property (Section~\ref{sec:exp-sweep}): on motion-heavy benchmarks, neither Diff Compression with a strong full-3D encoder nor TT3D with a mask-and-reconstruct objective unlocks the regime that the pair does; on appearance-discriminable benchmarks the non-leading cases bound the motion-priority interpretation (Appendix~\ref{sec:appendix-app-motion}).


\providecommand{\texbf}[1]{\textbf{#1}}

\section{Experiments}
\label{sec:experiments}

\subsection{Setup}
\label{sec:exp-setup}
All video SSL entries are evaluated under a shared small-scale pretraining and probing recipe. Entries that use an image-pretrained 2D path (TT3D, TT1D, and DisMo) initialize the spatial encoder from DINOv3 ViT-B/16 \citep{simeoni2025dinov3}; all parameters are trained jointly under the matched recipe, and partial-unfreezing variants are out of scope. This protocol is designed to compare architecture-objective choices while controlling data, schedule, parameter scale, and downstream evaluation. The claim we test is empirical: under the same shared video pretraining recipe, which combination yields a motion-prioritized representation? We use motion- and appearance-oriented dataset descriptions as empirical shorthand; Appendix~\ref{sec:appendix-app-motion} gives the quick diagnostic behind this interpretation.

Pretraining uses OpenVid \citep{DBLP:conf/iclr/NanXZFYCL0T25} at approximately 1M clips and Moments-in-Time v2 \citep{DBLP:journals/pami/MonfortVOAZRBYB20} at approximately 700k clips. We sample 8 frames at 6 fps. The default recipe is 8 epochs, effective global batch size 32, AdamW with peak learning rate 5e-4, betas $(0.9,0.98)$, weight decay $0.01$, gradient clipping at $0.1$, 10k linear warmup, cosine decay to $1\%$ of the peak learning rate, and fp16 mixed precision. We use $\mu$P \citep{yang2021tuning} with base dimension 256. The sweep covers four encoders: the VideoMAE-style joint space-time 3D ViT \citep{tong2022videomae} (henceforth ViT3D), DisMo-style 2D+3D, TT1D, and TT3D, at a roughly 170M$\sim$190M encoder scale. It also covers six objectives: MAE, Adaptive AR, naive AR, two-jump AR, MAE-Diff, and Diff Compression. MAE and Adaptive AR are established prior-work objectives; Diff Compression is the proposed objective, and naive AR, two-jump AR, and MAE-Diff are ablation rows that vary autoregressive horizon, multi-step prediction, and the addition of a diffusion head, respectively. In total, each entry sees approximately 13.6M video samples over approximately 425k optimizer steps.

The decoder has three initialization regimes. No pretraining uses a random decoder with regression or diffusion auxiliary loss. ImageNet pretraining uses ImageNet-1k \citep{deng2009imagenet} and the DINOv3 ViT-B class token as diffusion guidance to reconstruct the full image; cross-attention is not trained in this stage \citep{simeoni2025dinov3}. Video pretraining samples two frames from a training video, gives the decoder the later frame's class token and the earlier frame's spatial features as cross-attention guidance, and reconstructs the later frame. All pretrained decoders use effective global batch size 256 for 3 epochs.

The V-JEPA~2 row is trained at the same scale/recipe using its native 2+8 schedule. Since it has no per-frame cls-token, we mean-pool per-frame tokens as the motion embedding. The sweep uses no augmentation as a uniform constraint. This is a controlled comparison, although not a neutral one in every respect: DisMo's native training scheme uses motion-preserved augmentations, so no augmentation can handicap non-TT3D objectives. We therefore report DisMo's dual-augmentation ablation in \S\ref{sec:exp-sweep}.

\textbf{Benchmarks and scope.} Jester and Something-Something\,V2 define their classes by the movement itself \citep{materzynska2019jester, goyal2017something}, Diving48 separates dives only by body dynamics \citep{li2018resound}, and ARID's low light makes appearance unreliable \citep{xu2020arid}. HMDB51, IARD, and EPIC-Kitchens sit on the appearance side. \S\ref{sec:exp-analysis} then measures motion sensitivity per clip rather than per dataset. The 8-epoch budget is itself a design choice matched to DisMo's published scale, so learning speed under the shared recipe is a measured property of each pair rather than a confounder, and all conclusions are stated for this matched budget.

\subsection{Synergy: Architecture-Objective Interaction}
\label{sec:exp-sweep}
We sweep all four architectures and six objectives under the shared video pretraining recipe, evaluated with attentive probing on Jester, Something-Something V2 (SSv2), and ARID. Table~\ref{tab:sweep} reports the no-augmentation sweep with a common \texttt{decS\_imgnet} decoder and a single run per cell. Three cells in this grid coincide with canonical configurations from prior or proposed work: ViT3D+MAE recovers standard VideoMAE \citep{tong2022videomae}, DisMo+AdaAR recovers standard DisMo \citep{resslerantal2025dismo}, and TT1D/TT3D+DiffComp are the configurations we propose. We shade these cells in Table~\ref{tab:sweep} so they can be located at a glance.

\begin{table}[!ht]\centering
\caption{Pretraining sweep: top-1 classification accuracy (\%) via frozen attentive probing on Jester, Something-Something\,V2, and ARID. Each cell is one architecture-objective configuration under the matched recipe (\S\ref{sec:exp-setup}) with a common \texttt{decS\_imgnet} decoder, single run. \textbf{Bold} = best in block; \underline{underline} = second-best. \colorbox{vmaecolor}{Blue} = VideoMAE, \colorbox{dismocolor}{green} = DisMo (second number with dual augmentation), \colorbox{ourscolor}{orange} = ours.}
\label{tab:sweep}
\small
\begin{tabular}{llrrrrrr}
\toprule
 & & \multicolumn{6}{c}{Objective} \\
\cmidrule(lr){3-8}
Dataset & Arch & MAE & AdaAR & tjAR & AR & MAE-Diff & DiffComp \\
\midrule
Jester & ViT3D & \cellcolor{vmaecolor}39.47 & 12.85 & 13.01 & 12.85 & 31.68 & 19.82 \\
 & DisMo & 29.71 & \cellcolor{dismocolor}16.46 / \underline{46.95} & 15.94 & 13.54 & 21.66 & 12.98 \\
 & TT1D & 26.65 & 11.70 & 10.45 & 11.86 & 15.68 & \cellcolor{ourscolor}24.32 \\
 & TT3D & 11.07 & 12.85 & 10.42 & 11.83 & 10.32 & \cellcolor{ourscolor}\textbf{53.89} \\
\midrule
SSv2 & ViT3D & \cellcolor{vmaecolor}\underline{14.95} & 3.99 & 4.86 & 3.93 & 11.73 & 6.57 \\
 & DisMo & 11.20 & \cellcolor{dismocolor}5.29 / 13.33 & 4.35 & 3.54 & 7.89 & 5.53 \\
 & TT1D & 10.05 & 4.71 & 2.53 & 4.86 & 6.00 & \cellcolor{ourscolor}7.91 \\
 & TT3D & 4.98 & 2.26 & 4.16 & 4.39 & 4.82 & \cellcolor{ourscolor}\textbf{18.28} \\
\midrule
ARID & ViT3D & \cellcolor{vmaecolor}\textbf{24.37} & 13.58 & 15.66 & 14.12 & 18.75 & 17.70 \\
 & DisMo & 18.15 & \cellcolor{dismocolor}15.46 / \underline{22.45} & 14.87 & 14.02 & 15.71 & 16.51 \\
 & TT1D & 17.65 & 13.03 & 12.73 & 13.38 & 14.52 & \cellcolor{ourscolor}16.41 \\
 & TT3D & 11.49 & 17.65 & 13.92 & 15.76 & 7.71 & \cellcolor{ourscolor}21.98 \\
\bottomrule\end{tabular}
\end{table}

The successful region is sparse. Of the 24 architecture-objective cells, only ViT3D+MAE and TT3D+Diff Compression reach a high tier on the motion-focused Jester and SSv2 benchmarks in the no-augmentation condition. ViT3D+MAE gives 39.47 on Jester and 14.95 on SSv2. TT3D+Diff Compression gives 53.89 and 18.28, the highest cells for both datasets. With TT3D fixed, replacing Diff Compression by MAE drops Jester from 53.89 to 11.07 and SSv2 from 18.28 to 4.98. With Diff Compression fixed, replacing TT3D by ViT3D or DisMo gives 19.82 or 12.98 on Jester and 6.57 or 5.53 on SSv2. This paired drop indicates an interaction effect between TT3D and Diff Compression, not a generic effect of either component alone.

ARID is more mixed in the base sweep. ViT3D+MAE leads the ARID block at 24.37, while TT3D+Diff Compression reaches 21.98. After the decoder ablation selects the decoder S, video-pretrained setting, TT-VidT reaches 37.47 on ARID in the final comparison. The final method leads the selected final-comparison row by +13.1 on ARID, compared with VideoMAE at 24.37; +26.3 on Jester, compared with DisMo with dual augmentation at $46.95$; and +11.0 on SSv2, compared with VideoMAE at 14.95. These gaps summarize the selected final-comparison rows reported in Table~\ref{tab:final}. They describe the operating profile of TT-VidT under the matched recipe, not a universal ranking of architectures or objectives.

The non-leading cases bound the claim. On IARD, TT3D's lead does not hold: a ViT3D+DiffComp configuration reaches 77.03, compared with 67.03 for TT3D. IARD rewards retention of per-frame appearance features such as identity, clothing, and viewpoint context. On HMDB51, VideoMAE leads at 27.73 in the final table. Appendix~\ref{sec:appendix-app-motion} provides a quick appearance-vs-motion diagnostic that helps interpret these boundary cases.

DisMo needs its native encoder/decoder dual augmentation \citep{resslerantal2025dismo}. Enabling this scheme in our controlled recipe improves DisMo substantially: Jester $16.46\to46.95$, SSv2 $5.29\to13.33$, and ARID $15.46\to22.45$ with attentive probing. We call this entry DisMo with dual augmentation in the final comparison. Even with dual augmentation, it trails TT-VidT by 26.3 on Jester and 12.6 on SSv2.

\subsection{Decoder Ablation}
\label{sec:exp-decoder}
We ablate the DiT decoder \citep{peebles2023dit} with the encoder fixed to TT3D+Diff Compression. Table~\ref{tab:decoder} varies decoder size at three settings (S, B, L) and initialization source across random, ImageNet-pretrained, and video-pretrained forms.

\begin{table}[!ht]\centering
\caption{Decoder ablation on TT3D + Diff Compression (no augmentation). Rows: decoder size $\times$ init source (\texttt{rand} = random init with regression/diffusion auxiliary loss; \texttt{imgnet}/\texttt{video} = DiT decoder pretrained on ImageNet-1k or our video task). Cells: top-1 classification accuracy (\%) via frozen attentive probing (D48 = Diving48; EK-V = EPIC-Kitchens-100 verb, trimmed). Best per column \textbf{bold}, second \underline{underlined}.}
\label{tab:decoder}
\small\begin{adjustbox}{max width=\textwidth}
\begin{tabular}{llrrrrrrr}
\toprule
Size & Init & HMDB & ARID & IARD & Jester & SSv2 & D48 & EK-V \\
\midrule
S & rand+reg & 4.57 & 13.28 & 71.76 & 20.21 & 7.53 & 5.53 & 24.89 \\
 & rand+diff & 7.61 & 13.18 & \underline{79.01} & 28.46 & 11.71 & 6.70 & 23.88 \\
 & imgnet & 12.64 & 21.98 & 67.03 & 53.89 & 18.28 & 5.63 & 28.23 \\
 & video & \textbf{25.15} & \underline{37.47} & 74.84 & \textbf{73.25} & \textbf{25.92} & \underline{8.17} & \underline{32.54} \\
\midrule
B & imgnet & 9.20 & 19.99 & \textbf{83.19} & 61.00 & 17.00 & 6.09 & 26.56 \\
 & video & \underline{23.16} & \textbf{37.84} & 77.36 & \underline{69.99} & \underline{23.99} & \textbf{9.64} & \textbf{32.82} \\
\midrule
L & imgnet & 6.15 & 14.52 & 76.37 & 26.95 & 8.84 & 5.69 & 24.99 \\
 & video & 11.98 & 17.90 & 74.73 & 34.37 & 12.15 & 7.16 & 27.52 \\
\bottomrule\end{tabular}\end{adjustbox}
\end{table}

Pretraining matters even after 3 epochs. At size S, where all initialization sources are measured, video pretraining dominates ImageNet pretraining, and ImageNet pretraining dominates random initialization on motion-focused columns. On Jester, decoder S, video-pretrained reaches 73.25, compared with decoder S, ImageNet-pretrained at 53.89, decoder S, random init with diffusion auxiliary loss at 28.46, and decoder S, random init with regression auxiliary loss at 20.21. On SSv2, the same ordering is 25.92, 18.28, 11.71, and 7.53. Thus the decoder is not merely an output head; its pretraining changes the pressure placed on the encoder.

Larger is not always better. With video pretraining, decoder S reaches 73.25 on Jester and 25.92 on SSv2, while decoder L drops to 34.37 and 12.15. L therefore underperforms S by approximately 50\%$\sim$53\% on these motion-focused benchmarks. One plausible interpretation is that a larger decoder reduces pressure on the compact motion tokens, allowing more of the reconstruction burden to shift into decoder capacity; the motion embedding then carries less information overall, including less motion. We treat this as an empirical pattern observed in Table~\ref{tab:decoder}, consistent with non-monotonic decoder behavior reported for masked autoencoding in VideoMAE v2 \citep{wang2023videomaev2}.

The S/B comparison is scope-dependent. On video-pretrained decoders, decoder S and decoder B are comparable on the main motion-heavy columns: Jester is 73.25 versus 69.99 and SSv2 is 25.92 versus 23.99. On ImageNet-pretrained decoders, the gap can reach 27\% on HMDB. We adopt decoder S, video-pretrained as the default decoder for TT-VidT, which reaches 25.15 on HMDB, 37.47 on ARID, 74.84 on IARD, 73.25 on Jester, and 25.92 on SSv2. The same pattern is consistent in extended evaluation: video-pretrained decoders dominate ImageNet-pretrained decoders on EK-V verb classification, EK-V anticipation, and Diving48 fine-tuning, with per-size breakdowns reported in the supplementary material.

\subsection{Efficiency: Encoder FLOPs}
\label{sec:exp-efficiency}
\begin{wraptable}{r}{0.35\textwidth}
\centering
\caption{FLOPs at $256^2$ For each model}
\label{tab:efficiency_flops}
\small
\begin{tabular}{lr}
\toprule
Model & FLOPs (GF) \\
\midrule
TT-VidT TT1D & 400.5 \\
TT-VidT TT3D & 456.1 \\
DisMo & 874.7 \\
V-JEPA~2 & 1009.9 \\
VideoMAE & 1012.6 \\
\bottomrule
\end{tabular}
\end{wraptable}
TT3D's Temporal Transfer Layer operates on a compressed per-frame token budget. A $4{\times}$ per-axis spatial downsample reduces spatial tokens from $N=256$ to $N'=16$ per frame; with $K=8$ motion tokens, each frame contributes $24$ tokens and the temporal-attention sequence has $192$ tokens instead of $2048$ for full spatial-temporal attention. Table~\ref{tab:efficiency_flops} reports the corresponding encoder-only analytical FLOPs: TT3D costs 456.1 GF, about half of DisMo-2D-3D and less than half of VideoMAE-3D under the same $T{=}8$, $256^2$ setting. V-JEPA~2 is listed for context with its native tube setup, $T{=}16$ raw frames to $8$ latent frames; all other entries use $T{=}8$.

\subsection{Motion Probe and Controls}
\label{sec:exp-analysis}

\begin{table}[!ht]\centering
\vspace{-0.8em}
\caption{Left: motion-inversion probe on SSv2, flip\,/\,stay (\%) per input transform, plus relative accuracy drop with shuffled frames. Right: attribution controls, frozen attentive probing.}
\label{tab:probe}
\small\begin{adjustbox}{max width=\textwidth}
\begin{tabular}{lcccr}
\toprule
Model & H-flip & V-flip & Reverse & Shuf.\ drop \\
\midrule
\textbf{TT-VidT (Ours)} & \textbf{81.2 / 0.6} & \textbf{59.5 / 0.9} & \textbf{71.4 / 1.0} & 50.9\% \\
VideoMAE-3D & 66.9 / 9.7 & 55.1 / 9.1 & 38.6 / 20.9 & 50.0\% \\
DisMo & 34.5 / 35.0 & 8.4 / 43.2 & 34.5 / 16.4 & 31.9\% \\
DINOv3-1f & 37.1 / 24.5 & 11.8 / 22.7 & 0.0 / 100.0 & 3.1\% \\
\bottomrule\end{tabular}
\hspace{1.5em}
\begin{tabular}{lrrr}
\toprule
Model & Jester & SSv2 & ARID \\
\midrule
ViT3D+MAE & 39.47 & 14.95 & 24.37 \\
\;\;+ DINOv3 init & 39.53 & 14.00 & 24.86 \\
TT3D expl.\ diff & 64.7 & 24.8 & 26.5 \\
\textbf{TT-VidT (Ours)} & \textbf{73.25} & \textbf{25.92} & \textbf{37.47} \\
\bottomrule\end{tabular}\end{adjustbox}
\end{table}

Table~\ref{tab:probe} asks whether the motion-prioritized profile of \S\ref{sec:exp-sweep} lives in the representation itself, and which ingredient of TT-VidT produces it. The left half inverts the motion of SSv2 clips whose classes are exact mirrors under horizontal flip, vertical flip, or time reversal, re-encodes them, and asks an attentive probe trained on clean clips whether its decision follows (Appendix~\ref{sec:appendix-review}). A prediction that moves to the mirror class shows that the encoder re-read the motion, and one that stays shows that appearance decided it, since the content is otherwise unchanged. TT-VidT follows the inverted motion on nearly every clip, whereas every baseline keeps a share of its original answers, most of all under time reversal, the one transform that leaves every pixel untouched. The single-frame DINOv3 control cannot see frame order and never follows a reversal, which validates the probe and shows that Temporal Transfer and Diff Compression turn an order-blind substrate into the most direction-faithful encoder of the comparison. The shuffle column separates using frame order from reading it: VideoMAE depends on order as much as TT-VidT, yet reads its direction far less faithfully. The benchmark profile therefore reflects what the representation encodes rather than which datasets happen to reward it.

The right half isolates the two ingredients TT-VidT shares with other work. A ViT3D initialized from DINOv3 in the 12 of its 24 layers that the teacher can fill lands on its scratch counterpart on every motion-heavy benchmark and moves only the appearance-heavy IARD, so the image substrate shapes appearance rather than the motion regime. A VTok-style motion token computed as an explicit feature difference \citep{wang2026vtokunifiedvideotokenizer} already surpasses every non-TT configuration on Jester, which credits the keyframe factorization, and the learned token of TT-VidT improves on it on all five datasets, most where subtraction discards appearance that the task still needs. Neither the initialization nor the factorization alone explains the gain, which again locates it in the pairing of Temporal Transfer and Diff Compression. The gain also exceeds seed variance, since the weakest of nine TT-VidT measurements stays above the strongest ViT3D+MAE measurement on every motion-heavy benchmark, and it persists at a near-doubled budget, where TT-VidT has already saturated while the gap to the baseline stays wide (Appendix~\ref{sec:appendix-review}). Under the matched recipe, the motion advantage is thus a property of the method rather than of training speed.

\subsection{Final Comparison}
\label{sec:exp-comparison}
We compare TT-VidT against the entries identified by the sweep and ablations: VideoMAE, DisMo with dual augmentation, and an internally retrained V-JEPA~2 under the shared video pretraining recipe. Table~\ref{tab:final} reports frozen attentive probing on HMDB, ARID, IARD, Jester, SSv2, and EK-V verb classification, plus Diving48 full fine-tuning and EK-V anticipation.

\begin{table}[!ht]\centering
\caption{Final comparison: V-JEPA2 trained at the same scale/recipe plus selected entries from the sweep and ablations. Frozen attentive probe on the five sweep datasets and EK-V (verb classification, trimmed clips); D48~FT is full finetuning (50 ep); EK-V~Antic. is verb anticipation (frozen attentive on untrimmed observation windows). Best per column \textbf{bold}, second \underline{underlined}.}
\label{tab:final}
\small\begin{adjustbox}{max width=\textwidth}
\begin{tabular}{lrrrrrrrr}
\toprule
Method & HMDB & ARID & IARD & Jester & SSv2 & EK-V & D48 FT & EK-V Antic. \\
\midrule
V-JEPA2 (24L$\times$768d, 16f@12fps, tube=2) & 22.70 & 16.01 & 79.34 & 27.41 & 10.59 & \underline{32.61} & 8.02 & \textbf{23.07} \\
VideoMAE (768d, 24L) & \textbf{27.73} & \underline{24.37} & \underline{80.44} & 39.47 & \underline{14.95} & \textbf{35.62} & \underline{8.43} & \underline{22.81} \\
DisMo (768d, 12s+12t) & 22.57 & 22.45 & \textbf{89.89} & \underline{46.95} & 13.33 & 31.97 & 8.12 & 22.72 \\
\textbf{TT-VidT TT3D(Ours)} & \underline{25.15} & \textbf{37.47} & 74.84 & \textbf{73.25} & \textbf{25.92} & 32.54 & \textbf{18.63} & 21.33 \\
\bottomrule\end{tabular}\end{adjustbox}
\end{table}

The four final-comparison rows expose different operating regimes.

\begin{wraptable}{r}{0.34\textwidth}
\vspace{-1.1em}
\centering
\caption{Frozen probe vs.\ 30-epoch end-to-end finetuning.}
\label{tab:e2e}
\small\setlength{\tabcolsep}{4pt}
\begin{tabular}{lrr}
\toprule
 & Frozen & FT \\
\midrule
\multicolumn{3}{l}{\emph{Jester}} \\
ViT3D+MAE & 39.47 & 51.50 \\
DisMo dual-aug & 46.95 & 51.76 \\
\textbf{TT-VidT} & \textbf{73.25} & \textbf{72.79} \\
\midrule
\multicolumn{3}{l}{\emph{SSv2}} \\
ViT3D+MAE & 14.95 & 13.40 \\
DisMo dual-aug & 13.33 & 14.85 \\
\textbf{TT-VidT} & \textbf{25.92} & \textbf{27.72} \\
\bottomrule
\end{tabular}
\end{wraptable}

\textbf{TT-VidT is the only method in our comparison group to lead Jester, Something-Something\,V2, ARID, and Diving48 fine-tuning simultaneously}, with the largest gains concentrated on these motion-heavy evaluations: $37.47$ on ARID, $73.25$ on Jester, $25.92$ on Something-Something V2, and $18.63$ on Diving48 fine-tuning. These four datasets are where motion features add the most value relative to a single-frame appearance baseline.

\textbf{TT-VidT's lead survives end-to-end finetuning.} Table~\ref{tab:e2e} finetunes the full encoder on Jester and SSv2 for 30 epochs under one shared recipe with identical learning rates for all rows. TT-VidT stays 21.0 and 12.9 points ahead of the strongest baseline. The baselines need 12 to 35 points of supervised correction yet converge near 51.5 on Jester, below TT-VidT's frozen 73.25, while TT-VidT barely moves from its frozen probe, so supervised adaptation saturates the baselines rather than closing the gap.

\textbf{VideoMAE is the broadest baseline under this recipe.} It leads HMDB51 ($27.73$) and EPIC-Kitchens verb classification ($35.62$), and remains competitive on several motion-heavy datasets. This breadth, combined with its strong appearance retention, makes it the most reasonable single fallback when the downstream task mixes motion and static cues.

\textbf{DisMo improves substantially when its native dual-augmentation training is restored.} Without dual augmentation, DisMo+AdaAR is not competitive in the no-augmentation sweep; with dual augmentation, the same architecture leads IARD ($89.89$) and reaches competitive scores on Jester and SSv2. This indicates that DisMo's augmentation design is integral to its architecture rather than incidental.

\textbf{V-JEPA~2 trails most evaluations under the matched-recipe small-scale setting}, indicating that latent prediction at this small scale does not match its native large-scale operating point on motion-heavy or appearance-discriminable benchmarks. Its native large-scale recipe is outside our protocol. It does, however, lead EPIC-Kitchens verb anticipation ($23.07$), indicating that latent prediction retains an advantage on long-horizon prediction even under the matched recipe.

The mixed tasks expose the boundary. On EK-V verb classification, VideoMAE leads at 35.62 because verb recognition still benefits from object and scene context; TT-VidT reaches 32.54, DisMo 31.97, and V-JEPA~2 32.34. On EK-V anticipation, V-JEPA~2 reaches 23.07, compared with VideoMAE at 22.81, DisMo at 22.72, and TT-VidT at 21.33. Anticipation under untrimmed observation windows is a setting where V-JEPA~2's temporal prediction can show, even though the same row is weak on Jester and SSv2. HMDB51 and IARD follow the appearance-sensitive diagnostic in Appendix~\ref{sec:appendix-app-motion}: VideoMAE leads HMDB at 27.73, and DisMo leads IARD at 89.89. On Jester and SSv2, TT-VidT's lead holds across all six probe configurations: kNN, linear and MLP over mean-pooled or learned-weight features, and attentive probing. The supplementary material provides per-probe breakdowns for EK-V verb classification, EK-V anticipation, and Diving48 fine-tuning.


\section{Conclusion}
\label{sec:conclusion}



We presented TT-VidT, a video self-supervised learning design that decouples appearance and temporal change during pretraining. TT-VidT keeps static appearance in a wide first-frame spatial anchor and routes frame-specific change through a compact Temporal Transfer pathway. Diff Compression makes this bottleneck explicit: target frames are reconstructed from the anchor and motion tokens, so the useful information in the motion tokens is precisely what cannot be recovered from the anchor alone.

We tested this design with a controlled $4 \times 6$ architecture-objective sweep under a shared data, schedule, and encoder-scale recipe. The main empirical finding is that the effect comes from the pairing, not from either component independently: TT3D with MAE does not enter the strong-motion regime, and Diff Compression with ViT3D or DisMo encoders remains far below TT-VidT on Jester and Something-Something\,V2. After the decoder ablation selects the video-pretrained decS setting, TT-VidT is the only final-comparison method to lead Jester, Something-Something\,V2, ARID, and Diving48 fine-tuning simultaneously.

This explains why TT-VidT is strong on motion-sensitive evaluations: the narrow temporal path makes frame-to-frame change cheap and salient while the wide anchor still supplies appearance for reconstruction. The boundary cases are consistent with the same profile: HMDB51, IARD, and EPIC-Kitchens can reward identity, object and scene context, or anticipation, where broader appearance-sensitive or predictive baselines remain preferable. The same decoupling is also efficient: TT3D uses a 192-token temporal-attention sequence and 456.1 GF encoder FLOPs, roughly half of the parameter-comparable full video encoders in our accounting.

\paragraph{Limitations.}
This study is intentionally scoped to a small matched recipe: one pretraining mixture (approximately 1.7M OpenVid and Moments-in-Time v2 clips), single-run sweep cells with multi-seed checks on the canonical ones, and one roughly 170M$\sim$190M encoder scale. The resulting profile should therefore be read as evidence about this controlled regime rather than as a claim about all data scales or schedules. We also did not test combinations such as TT3D with DisMo-style dual augmentation, alternative appearance probes, or partial unfreezing of the spatial encoder.


\begin{ack}
This work is supported by the NVIDIA Taiwan AI Research \& Development Center (TRDC).
\end{ack}


\bibliographystyle{plainnat}
\bibliography{refs}

@inproceedings{peebles2023dit,
  title = {Scalable Diffusion Models with Transformers},
  author = {Peebles, William and Xie, Saining},
  booktitle = {Proceedings of the IEEE/CVF International Conference on Computer Vision},
  year = {2023},
  eprint = {2212.09748},
  archivePrefix = {arXiv},
  primaryClass = {cs.CV}
}

@inproceedings{tong2022videomae,
  title = {VideoMAE: Masked Autoencoders are Data-Efficient Learners for Self-Supervised Video Pre-Training},
  author = {Tong, Zhan and Song, Yibing and Wang, Jue and Wang, Limin},
  booktitle = {Advances in Neural Information Processing Systems},
  year = {2022}
}

@inproceedings{wang2023videomaev2,
  title = {VideoMAE V2: Scaling Video Masked Autoencoders with Dual Masking},
  author = {Wang, Limin and Huang, Bingkun and Zhao, Zhiyu and Tong, Zhan and He, Yinan and Wang, Yi and Wang, Yali and Qiao, Yu},
  booktitle = {Proceedings of the IEEE/CVF Conference on Computer Vision and Pattern Recognition},
  year = {2023}
}

@inproceedings{bardes2024vjepa,
  title = {Revisiting Feature Prediction for Learning Visual Representations from Video},
  author = {Bardes, Adrien and Garrido, Quentin and Ponce, Jean and Chen, Xinlei and Rabbat, Michael and LeCun, Yann and Assran, Mahmoud and Ballas, Nicolas},
  booktitle = {International Conference on Learning Representations},
  year = {2024}
}

@misc{assran2025vjepa2,
  title = {V-JEPA 2: Self-Supervised Video Models Enable Understanding, Prediction and Planning},
  author = {Assran, Mido and Bardes, Adrien and Fan, David and Garrido, Quentin and Howes, Russell and Komeili, Mojtaba and Muckley, Matthew and Rizvi, Ammar and Roberts, Claire and Sinha, Koustuv and Zholus, Artem and Arnaud, Sergio and Gejji, Abha and Martin, Ada and Hogan, Francois Robert and Dugas, Daniel and Bojanowski, Piotr and Khalidov, Vasil and Labatut, Patrick and Massa, Francisco and Szafraniec, Marc and Krishnakumar, Kapil and Li, Yong and Ma, Xiaodong and Chandar, Sarath and Meier, Franziska and LeCun, Yann and Rabbat, Michael and Ballas, Nicolas},
  year = {2025},
  eprint = {2506.09985},
  archivePrefix = {arXiv},
  primaryClass = {cs.AI}
}

@misc{ren2024arvideo,
  title = {ARVideo: Autoregressive Pretraining for Self-Supervised Video Representation Learning},
  author = {Ren, Sucheng and Zhu, Hongru and Wei, Chen and Li, Yijiang and Yuille, Alan and Xie, Cihang},
  year = {2024},
  eprint = {2405.15160},
  archivePrefix = {arXiv},
  primaryClass = {cs.CV}
}

@misc{rajasegaran2025toto,
  title = {An Empirical Study of Autoregressive Pre-training from Videos},
  author = {Rajasegaran, Jathushan and Radosavovic, Ilija and Ravishankar, Rahul and Gandelsman, Yossi and Feichtenhofer, Christoph and Malik, Jitendra},
  year = {2025},
  eprint = {2501.05453},
  archivePrefix = {arXiv},
  primaryClass = {cs.CV}
}

@misc{liu2025videomap,
  title = {VideoMAP: Toward Scalable Mamba-based Video Autoregressive Pretraining},
  author = {Liu, Yunze and Wu, Peiran and Liang, Cheng and Shen, Junxiao and Wang, Limin and Yi, Li},
  year = {2025},
  eprint = {2503.12332},
  archivePrefix = {arXiv},
  primaryClass = {cs.CV}
}

@misc{li2025nextvid,
  title = {Learning from Next-Frame Prediction: Autoregressive Video Modeling Encodes Effective Representations},
  author = {Li, Jinghan and Jin, Yang and Jiang, Hao and Mu, Yadong and Song, Yang and Xu, Kun},
  year = {2025},
  eprint = {2512.21004},
  archivePrefix = {arXiv},
  primaryClass = {cs.CV}
}

@inproceedings{arnab2021vivit,
  title = {ViViT: A Video Vision Transformer},
  author = {Arnab, Anurag and Dehghani, Mostafa and Heigold, Georg and Sun, Chen and Lucic, Mario and Schmid, Cordelia},
  booktitle = {Proceedings of the IEEE/CVF International Conference on Computer Vision},
  year = {2021}
}

@inproceedings{bertasius2021timesformer,
  title = {Is Space-Time Attention All You Need for Video Understanding?},
  author = {Bertasius, Gedas and Wang, Heng and Torresani, Lorenzo},
  booktitle = {International Conference on Machine Learning},
  year = {2021}
}

@inproceedings{pan2022stadapter,
  title = {ST-Adapter: Parameter-Efficient Image-to-Video Transfer Learning},
  author = {Pan, Junting and Lin, Ziyi and Zhu, Xiatian and Shao, Jing and Li, Hongsheng},
  booktitle = {Advances in Neural Information Processing Systems},
  year = {2022}
}

@inproceedings{yang2023aim,
  title = {AIM: Adapting Image Models for Efficient Video Action Recognition},
  author = {Yang, Taojiannan and Zhu, Yi and Xie, Yusheng and Zhang, Aston and Chen, Chen and Li, Mu},
  booktitle = {International Conference on Learning Representations},
  year = {2023}
}

@misc{simeoni2025dinov3,
  title = {DINOv3},
  author = {Sim{\'e}oni, Oriane and Vo, Huy V. and Seitzer, Maximilian and Baldassarre, Federico and Oquab, Maxime and Jose, Cijo and Khalidov, Vasil and Szafraniec, Marc and Yi, Seungeun and Ramamonjisoa, Micha{\"e}l and Massa, Francisco and Haziza, Daniel and Wehrstedt, Luca and Wang, Jianyuan and Darcet, Timoth{\'e}e and Moutakanni, Th{\'e}o and Sentana, Leonel and Roberts, Claire and Vedaldi, Andrea and Tolan, Jamie and Brandt, John and Couprie, Camille and Mairal, Julien and J{\'e}gou, Herv{\'e} and Labatut, Patrick and Bojanowski, Piotr},
  year = {2025},
  eprint = {2508.10104},
  archivePrefix = {arXiv},
  primaryClass = {cs.CV}
}

@inproceedings{resslerantal2025dismo,
  title = {DisMo: Disentangled Motion Representations for Open-World Motion Transfer},
  author = {Ressler-Antal, Thomas and Fundel, Frank and Ben Alaya, Malek and Baumann, Stefan Andreas and Krause, Felix and Gui, Ming and Ommer, Bj{\"o}rn},
  booktitle = {Advances in Neural Information Processing Systems},
  year = {2025}
}

@misc{tishby2000informationbottleneck,
  title = {The Information Bottleneck Method},
  author = {Tishby, Naftali and Pereira, Fernando C. and Bialek, William},
  year = {2000},
  eprint = {physics/0004057},
  archivePrefix = {arXiv},
  primaryClass = {physics.data-an}
}

@inproceedings{alemi2017deepvib,
  title = {Deep Variational Information Bottleneck},
  author = {Alemi, Alexander A. and Fischer, Ian and Dillon, Joshua V. and Murphy, Kevin},
  booktitle = {International Conference on Learning Representations},
  year = {2017}
}

@inproceedings{achille2018informationdropout,
  title = {Information Dropout: Learning Optimal Representations Through Noisy Computation},
  author = {Achille, Alessandro and Soatto, Stefano},
  booktitle = {International Conference on Learning Representations},
  year = {2018}
}

@inproceedings{kuehne2011hmdb,
  title = {HMDB: A Large Video Database for Human Motion Recognition},
  author = {Kuehne, Hildegard and Jhuang, Hueihan and Garrote, Estibaliz and Poggio, Tomaso and Serre, Thomas},
  booktitle = {Proceedings of the IEEE International Conference on Computer Vision},
  year = {2011}
}

@inproceedings{liu2021noframe,
  title = {No Frame Left Behind: Full Video Action Recognition},
  author = {Liu, Xin and Pintea, Silvia L. and Nejadasl, Fatemeh Karimi and Booij, Olaf and van Gemert, Jan C.},
  booktitle = {Proceedings of the IEEE/CVF Conference on Computer Vision and Pattern Recognition},
  year = {2021}
}

@inproceedings{fioresi2025albar,
  title = {ALBAR: Adversarial Learning approach to mitigate Biases in Action Recognition},
  author = {Fioresi, Joseph and Dave, Ishan Rajendrakumar and Shah, Mubarak},
  booktitle = {International Conference on Learning Representations},
  year = {2025},
  eprint = {2502.00156},
  archivePrefix = {arXiv},
  primaryClass = {cs.CV}
}

@inproceedings{materzynska2019jester,
  title = {The Jester Dataset: A Large-Scale Video Dataset of Human Gestures},
  author = {Materzynska, Joanna and Berger, Guillaume and Bax, Ingo and Memisevic, Roland},
  booktitle = {Proceedings of the IEEE/CVF International Conference on Computer Vision Workshops},
  year = {2019}
}

@inproceedings{goyal2017something,
  title = {The Something Something Video Database for Learning and Evaluating Visual Common Sense},
  author = {Goyal, Raghav and Ebrahimi Kahou, Samira and Michalski, Vincent and Materzynska, Joanna and Westphal, Susanne and Kim, Heuna and Haenel, Valentin and Fruend, Ingo and Yianilos, Peter and Mueller-Freitag, Moritz and Hoppe, Florian and Thurau, Christian and Bax, Ingo and Memisevic, Roland},
  booktitle = {Proceedings of the IEEE International Conference on Computer Vision},
  year = {2017}
}

@inproceedings{li2018resound,
  title = {RESOUND: Towards Action Recognition without Representation Bias},
  author = {Li, Yingwei and Li, Yi and Vasconcelos, Nuno},
  booktitle = {Proceedings of the European Conference on Computer Vision},
  pages = {513--528},
  year = {2018}
}

@inproceedings{kowal2022staticdynamic,
  title = {A Deeper Dive Into What Deep Spatiotemporal Networks Encode: Quantifying Static vs. Dynamic Information},
  author = {Kowal, Matthew and Siam, Mennatullah and Islam, Md Amirul and Bruce, Neil D. B. and Wildes, Richard P. and Derpanis, Konstantinos G.},
  booktitle = {Proceedings of the IEEE/CVF Conference on Computer Vision and Pattern Recognition},
  year = {2022}
}

@misc{xu2020arid,
  title = {ARID: A New Dataset for Recognizing Action in the Dark},
  author = {Xu, Yuecong and Yang, Jianfei and Cao, Haozhi and Mao, Kezhi and Yin, Jianxiong and See, Simon},
  year = {2020},
  eprint = {2006.03876},
  archivePrefix = {arXiv},
  primaryClass = {cs.CV}
}

@misc{tacchetti2017iard,
  title = {Invariant Action Recognition Dataset},
  author = {Tacchetti, Andrea and Isik, Leyla and Poggio, Tomaso},
  year = {2017},
  note = {Center for Brains, Minds and Machines dataset}
}

@article{isik2017fast,
  title = {Fast, Invariant Representation for Human Action in the Visual System},
  author = {Isik, Leyla and Tacchetti, Andrea and Poggio, Tomaso},
  journal = {Journal of Neurophysiology},
  year = {2018}
}

@article{damen2022epickitchens100,
  title = {Rescaling Egocentric Vision: Collection, Pipeline and Challenges for EPIC-KITCHENS-100},
  author = {Damen, Dima and Doughty, Hazel and Farinella, Giovanni Maria and Furnari, Antonino and Kazakos, Evangelos and Ma, Jian and Moltisanti, Davide and Munro, Jonathan and Perrett, Toby and Price, Will and Wray, Michael},
  journal = {International Journal of Computer Vision},
  year = {2022}
}

@misc{openmmlab2024kinetics710,
  title = {Preparing {Kinetics-710}},
  author = {{OpenMMLab}},
  year = {2024},
  howpublished = {MMAction2 dataset documentation},
  note = {\url{https://mmaction2.readthedocs.io/en/latest/dataset_zoo/kinetics710.html}}
}

@inproceedings{chen2020simclr,
  title = {A Simple Framework for Contrastive Learning of Visual Representations},
  author = {Chen, Ting and Kornblith, Simon and Norouzi, Mohammad and Hinton, Geoffrey},
  booktitle = {International Conference on Machine Learning},
  year = {2020}
}

@inproceedings{he2022mae,
  title = {Masked Autoencoders Are Scalable Vision Learners},
  author = {He, Kaiming and Chen, Xinlei and Xie, Saining and Li, Yanghao and Doll{\'a}r, Piotr and Girshick, Ross},
  booktitle = {Proceedings of the IEEE/CVF Conference on Computer Vision and Pattern Recognition},
  year = {2022}
}

@inproceedings{caron2021dino,
  title = {Emerging Properties in Self-Supervised Vision Transformers},
  author = {Caron, Mathilde and Touvron, Hugo and Misra, Ishan and J{\'e}gou, Herv{\'e} and Mairal, Julien and Bojanowski, Piotr and Joulin, Armand},
  booktitle = {Proceedings of the IEEE/CVF International Conference on Computer Vision},
  year = {2021}
}

@misc{oquab2023dinov2,
  title = {DINOv2: Learning Robust Visual Features without Supervision},
  author = {Oquab, Maxime and Darcet, Timoth{\'e}e and Moutakanni, Th{\'e}o and Vo, Huy V. and Szafraniec, Marc and Khalidov, Vasil and Fernandez, Pierre and Haziza, Daniel and Massa, Francisco and El-Nouby, Alaaeldin and Assran, Mahmoud and Ballas, Nicolas and Galuba, Wojciech and Howes, Russell and Huang, Po-Yao and Li, Shang-Wen and Misra, Ishan and Rabbat, Michael and Sharma, Vasu and Synnaeve, Gabriel and Xu, Hu and J{\'e}gou, Herv{\'e} and Mairal, Julien and Labatut, Patrick and Joulin, Armand and Bojanowski, Piotr},
  year = {2023},
  eprint = {2304.07193},
  archivePrefix = {arXiv},
  primaryClass = {cs.CV}
}

@inproceedings{bandara2022adamae,
  title = {AdaMAE: Adaptive Masking for Efficient Spatiotemporal Learning with Masked Autoencoders},
  author = {Bandara, Wele Gedara Chaminda and Patel, Naman and Gholami, Ali and Nikkhah, Mehdi and Agrawal, Motilal and Patel, Vishal M.},
  booktitle = {Proceedings of the IEEE/CVF Conference on Computer Vision and Pattern Recognition},
  year = {2023},
  eprint = {2211.09120},
  archivePrefix = {arXiv},
  primaryClass = {cs.CV}
}

@inproceedings{wang2023mvd,
  title = {Masked Video Distillation: Rethinking Masked Feature Modeling for Self-Supervised Video Representation Learning},
  author = {Wang, Rui and Chen, Dongdong and Wu, Zuxuan and Chen, Yinpeng and Dai, Xiyang and Liu, Mengchen and Yuan, Lu and Jiang, Yu-Gang},
  booktitle = {Proceedings of the IEEE/CVF Conference on Computer Vision and Pattern Recognition},
  year = {2023}
}

@misc{song2022mam2,
  title = {It Takes Two: Masked Appearance-Motion Modeling for Self-supervised Video Transformer Pre-training},
  author = {Song, Yuxin and Yang, Min and Wu, Wenhao and He, Dongliang and Li, Fu and Wang, Jingdong},
  year = {2022},
  eprint = {2210.05234},
  archivePrefix = {arXiv},
  primaryClass = {cs.CV}
}

@misc{yang2022motionmae,
  title = {Self-supervised Video Representation Learning with Motion-Aware Masked Autoencoders},
  author = {Yang, Haosen and Huang, Deng and Wen, Bin and Wu, Jiannan and Yao, Hongxun and Jiang, Yi and Zhu, Xiatian and Yuan, Zehuan},
  year = {2022},
  eprint = {2210.04154},
  archivePrefix = {arXiv},
  primaryClass = {cs.CV}
}

@inproceedings{sun2023mme,
  title = {Masked Motion Encoding for Self-Supervised Video Representation Learning},
  author = {Sun, Xinyu and Chen, Peihao and Chen, Liangwei and Li, Changhao and Li, Thomas H. and Tan, Mingkui and Gan, Chuang},
  booktitle = {Proceedings of the IEEE/CVF Conference on Computer Vision and Pattern Recognition},
  year = {2023},
  eprint = {2210.06096},
  archivePrefix = {arXiv},
  primaryClass = {cs.CV}
}

@inproceedings{thoker2025smile,
  title = {SMILE: Infusing Spatial and Motion Semantics in Masked Video Learning},
  author = {Thoker, Fida Mohammad and Jiang, Letian and Zhao, Chen and Ghanem, Bernard},
  booktitle = {Proceedings of the IEEE/CVF Conference on Computer Vision and Pattern Recognition},
  year = {2025},
  eprint = {2504.00527},
  archivePrefix = {arXiv},
  primaryClass = {cs.CV}
}

@misc{vandeghen2026trackmae,
  title = {TrackMAE: Video Representation Learning via Track Mask and Predict},
  author = {Vandeghen, Renaud and Thoker, Fida Mohammad and Van Droogenbroeck, Marc and Ghanem, Bernard},
  year = {2026},
  eprint = {2603.27268},
  archivePrefix = {arXiv},
  primaryClass = {cs.CV}
}

@inproceedings{dave2024nomoreshortcuts,
  title = {No More Shortcuts: Realizing the Potential of Temporal Self-Supervision},
  author = {Dave, Ishan Rajendrakumar and Jenni, Simon and Shah, Mubarak},
  booktitle = {Proceedings of the AAAI Conference on Artificial Intelligence},
  year = {2024},
  eprint = {2312.13008},
  archivePrefix = {arXiv},
  primaryClass = {cs.CV}
}

@article{wu2025advise,
  title = {Advancing Video Self-Supervised Learning via Image Foundation Models},
  author = {Wu, Jingwei and Huang, Zhewei and Liu, Chang},
  journal = {Pattern Recognition Letters},
  volume = {192},
  pages = {22--28},
  year = {2025},
  doi = {10.1016/j.patrec.2025.03.015},
  eprint = {2505.19218},
  archivePrefix = {arXiv},
  primaryClass = {cs.CV}
}

@misc{tv2025frame,
  title = {FRAME: Pre-Training Video Feature Representations via Anticipation and Memory},
  author = {TV, Sethuraman and Khosla, Savya and Srinivasakumar, Vignesh and Huang, Jiahui and Oh, Seoung Wug and Jenni, Simon and Hoiem, Derek and Lee, Joon-Young},
  year = {2025},
  eprint = {2506.05543},
  archivePrefix = {arXiv},
  primaryClass = {cs.CV}
}

@misc{li2025salt,
  title = {Rethinking JEPA: Compute-Efficient Video SSL with Frozen Teachers},
  author = {Li, Xianhang and Huang, Chen and Li, Chun-Liang and Malach, Eran and Susskind, Josh and Thilak, Vimal and Littwin, Etai},
  year = {2025},
  eprint = {2509.24317},
  archivePrefix = {arXiv},
  primaryClass = {cs.CV}
}

@inproceedings{qing2023dist,
  title = {Disentangling Spatial and Temporal Learning for Efficient Image-to-Video Transfer Learning},
  author = {Qing, Zhiwu and Zhang, Shiwei and Huang, Ziyuan and Zhang, Yingya and Gao, Changxin and Zhao, Deli and Sang, Nong},
  booktitle = {Proceedings of the IEEE/CVF International Conference on Computer Vision},
  year = {2023},
  eprint = {2309.07911},
  archivePrefix = {arXiv},
  primaryClass = {cs.CV}
}

@misc{li2024zeroi2v,
  title = {ZeroI2V: Zero-Cost Adaptation of Pre-trained Transformers from Image to Video},
  author = {Li, Xinhao and Zhu, Yuhan and Wang, Limin},
  year = {2024},
  eprint = {2310.01324},
  archivePrefix = {arXiv},
  primaryClass = {cs.CV}
}

@inproceedings{thoker2022severe,
  title = {How Severe is Benchmark-Sensitivity in Video Self-Supervised Learning?},
  author = {Thoker, Fida Mohammad and Doughty, Hazel and Bagad, Piyush and Snoek, Cees G. M.},
  booktitle = {European Conference on Computer Vision},
  year = {2022}
}

@misc{thoker2025severeplusplus,
  title = {SEVERE++: Evaluating Benchmark Sensitivity in Generalization of Video Representation Learning},
  author = {Thoker, Fida Mohammad and Jiang, Letian and Zhao, Chen and Bagad, Piyush and Doughty, Hazel and Ghanem, Bernard and Snoek, Cees G. M.},
  year = {2025},
  eprint = {2504.05706},
  archivePrefix = {arXiv},
  primaryClass = {cs.CV}
}

@misc{kumar2024benchmarkingvssl,
  title = {Benchmarking Self-Supervised Video Representation Learning},
  author = {Kumar, Akash and Kumar, Ashlesha and Vineet, Vibhav and Rawat, Yogesh S.},
  year = {2024},
  eprint = {2306.06010},
  archivePrefix = {arXiv},
  primaryClass = {cs.CV},
  note = {OpenReview, NeurIPS 2024 Datasets and Benchmarks Track submission}
}

@inproceedings{DBLP:conf/iclr/NanXZFYCL0T25, author = {Kepan Nan and Rui Xie and Penghao Zhou and Tiehan Fan and Zhenheng Yang and Zhijie Chen and Xiang Li and Jian Yang and Ying Tai}, title = {OpenVid-1M: {A} Large-Scale High-Quality Dataset for Text-to-video Generation}, booktitle = {The Thirteenth International Conference on Learning Representations, {ICLR} 2025, Singapore, April 24-28, 2025}, publisher = {OpenReview.net}, year = {2025}, url = {https://openreview.net/forum?id=j7kdXSrISM}, bibsource = {dblp computer science bibliography, https://dblp.org} }

@article{DBLP:journals/pami/MonfortVOAZRBYB20, author = {Mathew Monfort and Carl Vondrick and Aude Oliva and Alex Andonian and Bolei Zhou and Kandan Ramakrishnan and Sarah Adel Bargal and Tom Yan and Lisa M. Brown and Quanfu Fan and Dan Gutfreund}, title = {Moments in Time Dataset: One Million Videos for Event Understanding}, journal = {{IEEE} Trans. Pattern Anal. Mach. Intell.}, volume = {42}, number = {2}, pages = {502--508}, year = {2020}, url = {https://doi.org/10.1109/TPAMI.2019.2901464}, doi = {10.1109/TPAMI.2019.2901464}, bibsource = {dblp computer science bibliography, https://dblp.org} }

@inproceedings{deng2009imagenet,
  title={Imagenet: A large-scale hierarchical image database},
  author={Deng, Jia and Dong, Wei and Socher, Richard and Li, Li-Jia and Li, Kai and Fei-Fei, Li},
  booktitle={2009 IEEE conference on computer vision and pattern recognition},
  pages={248--255},
  year={2009},
  organization={IEEE}
}

@inproceedings{
yang2021tuning,
title={Tuning Large Neural Networks via Zero-Shot Hyperparameter Transfer},
author={Greg Yang and Edward J Hu and Igor Babuschkin and Szymon Sidor and Xiaodong Liu and David Farhi and Nick Ryder and Jakub Pachocki and Weizhu Chen and Jianfeng Gao},
booktitle={Advances in Neural Information Processing Systems},
editor={A. Beygelzimer and Y. Dauphin and P. Liang and J. Wortman Vaughan},
year={2021},
url={https://openreview.net/forum?id=Bx6qKuBM2AD}
}

@inproceedings{DBLP:conf/iclr/LipmanCBNL23, author = {Yaron Lipman and Ricky T. Q. Chen and Heli Ben{-}Hamu and Maximilian Nickel and Matthew Le}, title = {Flow Matching for Generative Modeling}, booktitle = {The Eleventh International Conference on Learning Representations, {ICLR} 2023, Kigali, Rwanda, May 1-5, 2023}, publisher = {OpenReview.net}, year = {2023}, url = {https://openreview.net/forum?id=PqvMRDCJT9t}, bibsource = {dblp computer science bibliography, https://dblp.org} }

@misc{wang2026vtokunifiedvideotokenizer,
      title={VTok: A Unified Video Tokenizer with Decoupled Spatial-Temporal Latents}, 
      author={Feng Wang and Yichun Shi and Ceyuan Yang and Qiushan Guo and Jingxiang Sun and Alan Yuille and Peng Wang},
      year={2026},
      eprint={2602.04202},
      archivePrefix={arXiv},
      primaryClass={cs.CV},
      url={https://arxiv.org/abs/2602.04202}, 
}


\clearpage
\newpage
\clearpage
\appendix
\renewcommand{\topfraction}{0.95}\renewcommand{\bottomfraction}{0.95}\renewcommand{\floatpagefraction}{0.9}
\setcounter{topnumber}{4}\setcounter{bottomnumber}{2}\setcounter{totalnumber}{6}
\setlength{\abovecaptionskip}{4pt plus 1pt minus 1pt}\setlength{\belowcaptionskip}{3pt plus 1pt minus 1pt}
\setlength{\textfloatsep}{10pt plus 2pt minus 2pt}\setlength{\floatsep}{8pt plus 2pt minus 2pt}
\setlength{\intextsep}{8pt plus 2pt minus 2pt}
\makeatletter\setlength{\@fptop}{0pt}\setlength{\@fpsep}{\floatsep}\setlength{\@fpbot}{0pt plus 1fil}\makeatother
\noindent\rule{\linewidth}{0.4pt}
\begin{center}
    {\Large\bfseries Appendix}
\end{center}
\vspace{-0.5em}
\noindent\rule{\linewidth}{0.4pt}

\section*{Table of Contents}
\definecolor{sectionblue}{RGB}{65, 105, 225}
\titlecontents{section}[1.6em]{\addvspace{0.15pc}\color{sectionblue}}{\contentslabel[\thecontentslabel]{1.75em}}{\hspace*{-1.5em}}{\hfill\contentspage}[]
\titlecontents{subsection}[4em]{\addvspace{0pc}\color{sectionblue}}{\contentslabel[\thecontentslabel]{2.5em}}{\hspace*{-2.5em}}{\titlerule*[1pc]{.}\contentspage}[]
\startcontents[sections]
{\footnotesize\printcontents[sections]{l}{1}{\setcounter{tocdepth}{2}}}
\clearpage

\section{Appearance-vs-Motion Diagnostic}
\label{sec:appendix-app-motion}

This section provides a lightweight diagnostic for interpreting where TT-VidT helps and where broader appearance-sensitive representations remain stronger. The experiment is intended as a quick observation and source of insight, not as a formal benchmark or proof of motion understanding.

\begin{wrapfigure}{r}{0.45\textwidth}
\centering
\includegraphics[width=0.42\textwidth]{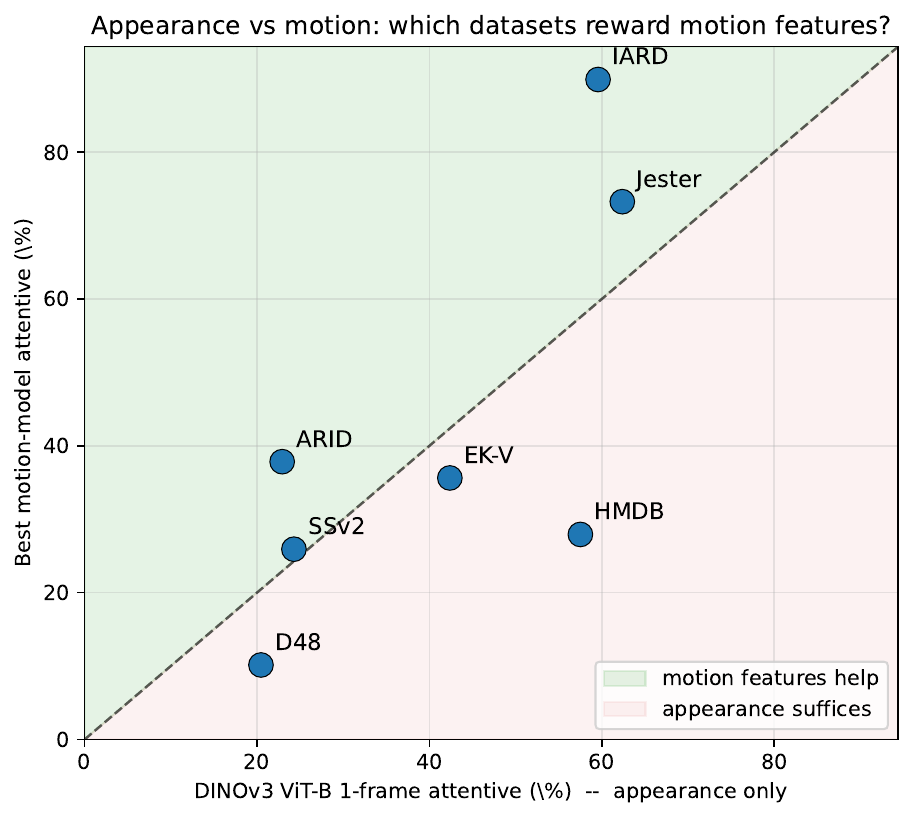}
\caption{Each point is a dataset. X-axis: DINOv3 ViT-B single-frame attentive accuracy (appearance baseline). Y-axis: best motion-model attentive across our entries. Region above the diagonal: motion features add value over appearance.}
\label{fig:app_vs_motion}
\end{wrapfigure}

We compare each dataset's best motion-aware score from our trained video entries against a single-frame DINOv3 ViT-B attentive baseline, used as an appearance-only reference \citep{simeoni2025dinov3}. Figure~\ref{fig:app_vs_motion} plots DINOv3 single-frame accuracy on the x-axis and the best motion-model attentive score on the y-axis. Points above the diagonal indicate datasets where temporal information appears to add value over single-frame recognition; points at or below the diagonal suggest appearance-dominated regimes. Since a single frame can still contain implicit motion cues, this comparison should be read as suggestive evidence rather than a strict separation of appearance and motion.

The diagnostic helps explain the profile observed in the main experiments. Jester and ARID lie above the diagonal: their best frozen attentive probe over trained video entries exceeds the single-frame DINOv3 baseline by a clear margin, consistent with TT-VidT's gains concentrating on motion-sensitive recognition. Something-Something V2 is not shown in the diagnostic plot but exhibits the same pattern in the architecture-objective sweep (Table~\ref{tab:sweep}). Diving48 lies below the diagonal in the frozen-probe view, but full fine-tuning recovers a strong gain ($18.63$ vs. DisMo with dual augmentation $8.12$); we therefore treat Diving48 as complementary fine-tuning evidence rather than a frozen-probe diagnostic point.

HMDB51 and IARD illustrate the opposite side of the profile. HMDB51 is appearance-discriminable: a single-frame DINOv3 probe already explains much of the recognition signal, leaving less room for motion features. IARD requires a separate identity-control check because its appearance-vs-motion position is dominated by actor identity rather than by motion modeling. Table~\ref{tab:iard_identity} reports frozen $k$NN@20 identity classification over the five actors in IARD, where lower identity accuracy indicates less retained per-frame identity.

\begin{table}[h]\centering\small
\caption{IARD identity classification (5 actors, random split). $k$NN@20 over frozen attentive features. \textbf{Lower is better:} a motion-prioritized representation should retain less per-frame identity. Random baseline is $20.00\%$.}
\label{tab:iard_identity}
\begin{tabular}{lr}
\toprule
Method & $k$NN@20 \\
\midrule
DINOv3 1f (appearance reference) & 100.00 \\
V-JEPA~2, 16f & 98.79 \\
DisMo with dual augmentation & 99.56 \\
VideoMAE & 98.90 \\
\textbf{TT-VidT} & \textbf{70.99} \\
random baseline & 20.00 \\
\bottomrule
\end{tabular}
\end{table}

Across the four pretrained video baselines, only TT-VidT's identity accuracy ($70.99$) drops substantially below the appearance reference ($100.00$); V-JEPA~2 ($98.79$), DisMo with dual augmentation ($99.56$), and VideoMAE ($98.90$) all retain near-complete identity information. This supports the interpretation that TT-VidT's compact motion channel discards more per-frame identity than the broader baselines. EPIC-Kitchens similarly mixes verb, noun, and scene signals, so broader appearance-sensitive baselines retain an advantage on its mixed evaluations. These observations bound the motion-priority claim: TT-VidT targets motion-sensitive recognition, while appearance-heavy or anticipation-heavy settings can favor other operating points.

\section{Architecture Sizing}
\label{sec:appendix-arch}

We report analytical parameter counts and forward-pass FLOPs for every encoder and decoder configuration used in the paper. Encoder FLOPs in Table~\ref{tab:encoder_flops} cover one full forward pass over a video of $T{=}8$ frames at $256{\times}256$, including the per-frame spatial path and any temporal mixing. The DINOv3 ViT-B row is reported both per-frame and at the $T{=}8$ scale, so it can be read against the temporal architectures on equal footing. V-JEPA2 ingests $T{=}16$ raw frames and produces $8$ latent frames through tube embedding; we list both numbers for transparency. TT-VidT in either TT1D or TT3D form remains within roughly $1.2\times$ of the $T{=}8$ per-frame DINOv3 cost, which is the basis for the matched comparison setting in the main paper.

Table~\ref{tab:decoder_flops} lists the three decoder sizes (S, B, L) used in the decoder ablation, with and without first-frame cross-attention. The xattn rows correspond to the appearance-anchored configuration used by Diff Compression or AR-series setup, where the decoder cross-attends to the full $z_k$ ($k=1$ for diff compression, $k=t-dt$ for AR series). The default decoder used throughout the main experiments is decS with cross-attention enabled.

\begin{table}[htbp]\scriptsize
\begin{minipage}[t]{0.52\linewidth}\centering
\caption{Encoder analytical parameters and forward-pass FLOPs at $T{=}8$, $256{\times}256$.}
\label{tab:encoder_flops}
\setlength{\tabcolsep}{3pt}
\begin{adjustbox}{max width=\linewidth}
\begin{tabular}{lrrl}
\toprule
Encoder & Params & FLOPs (GF) & Scope \\
\midrule
DINOv3 ViT-B (1-frame) & 85.5M & 47.1 & 1 frame, $256^2$ \\
DINOv3 ViT-B ($T{=}8$ per-frame) & 85.5M & 377.2 & $T{=}8$, $256^2$ \\
VideoMAE-3D (24L, 768d) & 170.5M & 1012.6 & $T{=}8$, $256^2$ \\
DisMo-2D-3D (12s+12t, 768d) & 170.5M & 874.7 & $T{=}8$, $256^2$ \\
TT-VidT TT1D (ours) & 175.2M & 400.5 & $T{=}8$, $256^2$ \\
TT-VidT TT3D (ours) & 196.4M & 456.1 & $T{=}8$, $256^2$ \\
V-JEPA2 (24L $\times$ 768d, tube=2) & 171.0M & 1009.9 & $T{=}16$ ($8$ latent), $256^2$ \\
\bottomrule
\end{tabular}
\end{adjustbox}
\end{minipage}
\hfill
\begin{minipage}[t]{0.46\linewidth}\centering
\caption{Decoder analytical parameters and forward-pass FLOPs at $T{=}8$. $L$, $D$, and $I$ denote depth, hidden dim, and FFN inner dim. xattn marks decoders with first-frame cross-attention.}
\label{tab:decoder_flops}
\setlength{\tabcolsep}{3pt}
\begin{adjustbox}{max width=\linewidth}
\begin{tabular}{lrrrcrr}
\toprule
Variant & $L$ & $D$ & $I$ & xattn & Params & FLOPs (GF) \\
\midrule
decS & 12 & 768 & 3072 & $-$ & 129.8M & 133.8 \\
decS & 12 & 768 & 3072 & \checkmark & 165.2M & 157.8 \\
decB & 16 & 1024 & 4096 & $-$ & 306.2M & 316.3 \\
decB & 16 & 1024 & 4096 & \checkmark & 390.1M & 373.2 \\
decL & 28 & 1152 & 3456 & $-$ & 562.7M & 584.1 \\
decL & 28 & 1152 & 3456 & \checkmark & 748.5M & 710.1 \\
\bottomrule
\end{tabular}
\end{adjustbox}
\end{minipage}
\end{table}

\section{Encoder Pretraining}
\label{sec:appendix-encoder-pretrain}

Encoder pretraining uses OpenVid-1M \citep{DBLP:conf/iclr/NanXZFYCL0T25} at 384 px together with Moments-in-Time \citep{DBLP:journals/pami/MonfortVOAZRBYB20}, sampled at 6 FPS to $T{=}8$ frames at $256{\times}256$. Frames are mapped to a $16{\times}16{\times}4$ latent through a frame VAE before reconstruction. The same recipe applies to every encoder in the architecture-objective sweep, so differences across rows of the main sweep table reflect architecture and objective rather than schedule. All encoder pretraining ran on B200 GPUs. Hyperparameters are listed in Table~\ref{tab:pretrain_hparams}.

\section{Decoder Pretraining}
\label{sec:appendix-decoder-pretrain}

The decoder is pretrained separately so that the same checkpoint can be reused across encoder runs in the sweep. We use either ImageNet-1k \citep{deng2009imagenet} (single-frame, repeated to fill the $T{=}33$ frame budget) or the same video corpus as the encoder, and evaluate both initializations in the main paper. For decoder-only pretraining, we use latent regression as a lightweight surrogate for the Diff Compression reconstruction target. Decoder pretraining ran on H100 GPUs, separate from the B200 hardware used for encoder pretraining and finetuning. Hyperparameters are in Table~\ref{tab:dec_pretrain_hparams}.

\begin{table}[htbp]\scriptsize
\begin{minipage}[t]{0.49\linewidth}\centering
\caption{Encoder pretraining hyperparameters. Hardware: B200.}
\label{tab:pretrain_hparams}
\setlength{\tabcolsep}{3pt}
\begin{adjustbox}{max width=\linewidth}
\begin{tabular}{ll}
\toprule
Setting & Value \\
\midrule
Pretraining datasets & OpenVid-1M (384 px) + Moments-in-Time v2 \\
Frame count $T$ & 8 \\
Sample FPS & 6 \\
Resolution & $256 \times 256$ \\
Latent space (frame VAE) & $16 \times 16 \times 4$ \\
Spatial backbone init & DINOv3 ViT-B/16 \\
Decoder & DiT, S (default) / B / L \\
Decoder pretrain init & ImageNet / video \\
Motion tokens $K$ & 8 \\
TT3D spatial downsample $f$ & 4 ($16{\times}16 \to 4{\times}4$) \\
Optimizer & AdamW \\
Betas & $(0.9, 0.98)$ \\
Weight decay & 0.01 \\
Peak learning rate & 5e-4 \\
LR schedule & cosine, min ratio 0.01 \\
Warmup steps & 10\,000 \\
Gradient clip & 0.1 \\
Epochs & 8 (main) / 20 (long-training) \\
Per-GPU batch size & 16 \\
GPUs & 2 B200 \\
Effective global batch & 32 (2 GPU $\times$ 16) \\
Gradient accumulation & 1 \\
\bottomrule
\end{tabular}
\end{adjustbox}
\end{minipage}
\hfill
\begin{minipage}[t]{0.49\linewidth}\centering
\caption{Decoder pretraining hyperparameters (DiT, S/B/L). Effective global batch $256$ across all three sizes. Hardware: H100.}
\label{tab:dec_pretrain_hparams}
\setlength{\tabcolsep}{3pt}
\begin{adjustbox}{max width=\linewidth}
\begin{tabular}{ll}
\toprule
Setting & Value \\
\midrule
Epochs & 3 \\
Effective global batch & 256 \\
Per-GPU batch & 8 \\
Gradient accumulation & 1 (S) / 2 (B) / 4 (L) \\
Optimizer & AdamW \\
Betas & $(0.9, 0.98)$ \\
Weight decay & 0.01 \\
Peak learning rate & 5e-4 \\
LR schedule & cosine, min ratio 0.01 \\
Warmup steps & 1\,000 \\
Frame count $T$ & 33 (multi-jump token counts $1/2/4/8$) \\
Resolution & $256 \times 256$ pixel ($16 \times 16$ latent) \\
Latent space & $16 \times 16 \times 4$ (frame VAE) \\
Decoder objective & latent regression (DiffComp-style) \\
Pretrain data, ImageNet init & ImageNet-1k (single frame, repeated) \\
Pretrain data, Video init & OpenVid-1M (384 px) + Moments-in-Time v2 \\
\bottomrule
\end{tabular}
\end{adjustbox}
\end{minipage}
\end{table}

\section{Frozen-Probe Evaluation}
\label{sec:appendix-probe}

Frozen-probe evaluation uses the protocol summarized in Table~\ref{tab:probe_hparams}. We report six probes (kNN, linear, layer-weighted linear, MLP, layer-weighted MLP, attentive) over mean-pooled features. Small datasets (HMDB51, ARID, IARD, Diving48) use batch $64$ for $100$ epochs; large datasets (Jester, SSv2, EK-100) use batch $256$ for $20$ epochs. Features are extracted at $T{=}8$ by default, with $T{=}16$ used for Diving48 where longer temporal context matters. We use ImageNet normalization for DINOv3 and V-JEPA2 baselines, and $[0.5, 0.5, 0.5]$ normalization for our models, matching the statistics each backbone was trained on.

\section{End-to-End Finetuning}
\label{sec:appendix-finetune}

We finetune end-to-end on Diving48 for 50 epochs. The encoder uses a much smaller learning rate than the head ($1\mathrm{e}{-}5$ vs $1\mathrm{e}{-}3$) to preserve the pretrained features while still allowing adaptation. The classification head is an attentive pool followed by a 2-layer MLP. We keep augmentation deliberately light (resize and horizontal flip on train; resize only on eval) so that the comparison reflects what the encoder offers rather than augmentation engineering. Finetuning ran on B200 GPUs. Hyperparameters are in Table~\ref{tab:ft_hparams}.

\begin{table}[htbp]\scriptsize
\begin{minipage}[t]{0.49\linewidth}\centering
\caption{Frozen-probe evaluation hyperparameters.}
\label{tab:probe_hparams}
\setlength{\tabcolsep}{3pt}
\begin{adjustbox}{max width=\linewidth}
\begin{tabular}{ll}
\toprule
Setting & Value \\
\midrule
Frozen probes & kNN, linear, linear-lw, MLP, MLP-lw, attentive (6) \\
kNN $k$ & 20 (cosine NN on mean-pooled features) \\
Probe MLP hidden & 512 \\
Probe MLP dropout & 0.1 \\
Probe optimizer & AdamW \\
Probe learning rate & 1e-3 \\
Probe weight decay & 1e-4 \\
Probe LR schedule & constant \\
Probe batch (small datasets) & 64 (HMDB51, ARID, IARD, Diving48) \\
Probe batch (large datasets) & 256 (Jester, SSv2, EK-100) \\
Probe epochs (small) & 100 \\
Probe epochs (large) & 20 \\
Feature extraction frames $T$ & 8 (default) / 16 (Diving48) \\
Sampling mode & center (eval), uniform (train) \\
Resolution & $256 \times 256$ \\
Normalization & ImageNet (DINOv3, V-JEPA2) or $[0.5, 0.5, 0.5]$ (ours) \\
\bottomrule
\end{tabular}
\end{adjustbox}
\end{minipage}
\hfill
\begin{minipage}[t]{0.49\linewidth}\centering
\caption{End-to-end finetuning hyperparameters. Hardware: B200.}
\label{tab:ft_hparams}
\setlength{\tabcolsep}{3pt}
\begin{adjustbox}{max width=\linewidth}
\begin{tabular}{ll}
\toprule
Setting & Value \\
\midrule
Finetune dataset & Diving48 (50 epochs) \\
Encoder LR & 1e-5 \\
Head LR & 1e-3 \\
Weight decay & 0.05 \\
Warmup epochs & 5 \\
Per-GPU batch & 2 \\
Gradient accumulation & 16 (effective batch 32) \\
Head & Attentive pool + 2-layer MLP (hidden 512, dropout 0.1) \\
Augmentation (train) & Resize 256, RandomHorizontalFlip, ImageNet norm \\
Augmentation (eval) & Resize 256, ImageNet norm \\
LR schedule & cosine \\
Optimizer & AdamW \\
\bottomrule
\end{tabular}
\end{adjustbox}
\end{minipage}
\end{table}

\section{Total Compute Footprint}
\label{sec:appendix-compute}

We report the wall-clock cost of every component of the work. Encoder pretraining and downstream finetuning ran on B200 GPUs; decoder pretraining ran on H100 GPUs. GPU-hours are reported on the hardware actually used for each component, and we do not attempt a cross-GPU normalization. The encoder pretrain row in Table~\ref{tab:compute_encoder_pretrain} reports the median run cost per architecture using decS, taken over the runs in the architecture-objective sweep. Table~\ref{tab:compute_decoder_pretrain} gives the per-checkpoint cost for the six decoder configurations (S/B/L $\times$ ImageNet/Video init), all at effective global batch $256$. Table~\ref{tab:compute_total} aggregates across all components for the full study.

\begin{table}[htbp]\scriptsize
\begin{minipage}[t]{0.22\linewidth}\centering
\caption{Encoder pretraining cost on B200, median across runs in the sweep, decS only.}
\label{tab:compute_encoder_pretrain}
\setlength{\tabcolsep}{3pt}
\begin{adjustbox}{max width=\linewidth}
\begin{tabular}{lr}
\toprule
Architecture & GPU-hours (B200) \\
\midrule
TT1D (ours)  & 78.5 \\
TT3D (ours)  & 87.7 \\
DisMo-2D-3D  & 91.1 \\
VideoMAE-3D  & 92.0 \\
\bottomrule
\end{tabular}
\end{adjustbox}
\end{minipage}
\hfill
\begin{minipage}[t]{0.40\linewidth}\centering
\caption{Decoder pretraining cost on H100, per checkpoint at effective global batch $256$.}
\label{tab:compute_decoder_pretrain}
\setlength{\tabcolsep}{3pt}
\begin{adjustbox}{max width=\linewidth}
\begin{tabular}{llrrr}
\toprule
Variant & Init & GPUs & Wall time & GPU-hours (H100) \\
\midrule
decS & ImageNet & 2 &  9 h 00 & 18.0 \\
decS & Video    & 2 & 12 h 20 & 24.7 \\
decB & ImageNet & 2 & 11 h 15 & 22.5 \\
decB & Video    & 2 & 15 h 20 & 30.7 \\
decL & ImageNet & 4 &  8 h 20 & 33.3 \\
decL & Video    & 4 & 11 h 50 & 47.3 \\
\bottomrule
\end{tabular}
\end{adjustbox}
\end{minipage}
\hfill
\begin{minipage}[t]{0.34\linewidth}\centering
\caption{Total compute footprint of this work. Decoder pretraining is on H100; all other components are on B200. We do not normalize across GPU types.}
\label{tab:compute_total}
\setlength{\tabcolsep}{3pt}
\begin{adjustbox}{max width=\linewidth}
\begin{tabular}{llr}
\toprule
Component & Hardware & GPU-hours \\
\midrule
Decoder pretrain (6 ckpts)            & H100 & $\approx 177$ \\
Encoder pretrain (decS, $n{=}35$)     & B200 & $3{,}663$ \\
Diving48 finetune ($\approx 30$ runs) & B200 & $\approx 700$ \\
Feature extraction                    & B200 & $24$ \\
Frozen probing                        & B200 & $19$ \\
\midrule
\textbf{Total}                        &      & \textbf{$\approx 4{,}580$} \\
\bottomrule
\end{tabular}
\end{adjustbox}
\end{minipage}
\end{table}

\section{Complete Results}
\label{sec:appendix-complete}

This section reports every pretrained run of this work on every evaluation. All runs follow the matched recipe of Section~\ref{sec:exp-setup} unless a table names a change. Frozen evaluation uses the six probes of Appendix~\ref{sec:appendix-probe}: $k$NN@20, linear and MLP heads over mean-pooled or learned-weight pooled tokens, and the attentive probe used in the main text. Finetuning follows Appendix~\ref{sec:appendix-finetune}.

\subsection{Architecture-Objective Sweep under Every Probe}
\label{sec:appendix-complete-sweep}

\begin{figure}[htbp]\centering
\includegraphics[width=\textwidth]{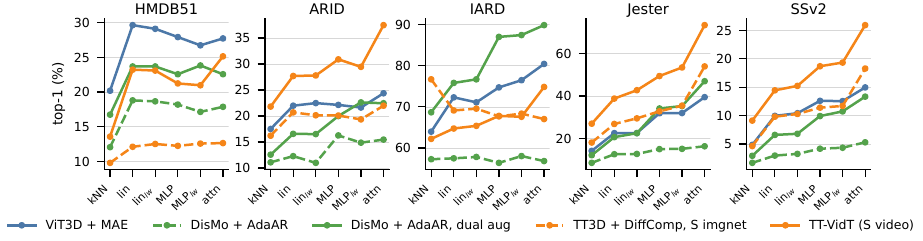}
\caption{The canonical configurations under the six frozen probes, from the weakest ($k$NN) to the strongest (attentive) readout.}
\label{fig:app_probe_profiles}
\end{figure}

\begin{figure}[htbp]\centering
\includegraphics[width=\textwidth]{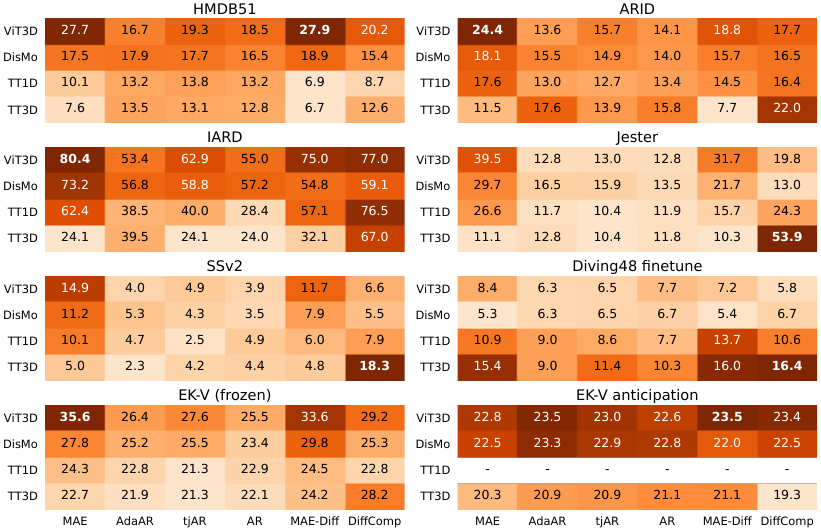}
\caption{Attentive probe (finetune for Diving48) of the 24 architecture-objective cells on every task. Colour is scaled per panel, the best cell is bold.}
\label{fig:app_sweep_heatmaps}
\end{figure}

Figure~\ref{fig:app_probe_profiles} follows the canonical cells across the six probes, and Figure~\ref{fig:app_sweep_heatmaps} shows the attentive probe of all 24 cells on every task. TT3D with Diff Compression is the darkest cell of its row on Jester, SSv2, ARID and Diving48 finetuning, and the lead of TT-VidT on Jester and SSv2 holds under every probe. Tables~\ref{tab:app_sweep_hmdb51_arid} to~\ref{tab:app_sweep_ekv_ekv_antic} give every cell under every probe, and Tables~\ref{tab:app_sweep_ft_d48} and~\ref{tab:app_sweep_ft_ekv} the end-to-end finetunes.

\begin{table}[htbp]\scriptsize
\begin{minipage}[t]{0.485\linewidth}\centering
\caption{End-to-end finetuning of every sweep cell on Diving48, 50 epochs.}
\label{tab:app_sweep_ft_d48}
\setlength{\tabcolsep}{3pt}\renewcommand{\arraystretch}{0.88}
\begin{adjustbox}{max width=\linewidth}
\begin{tabular}{lrrrrrr}
\toprule
Arch & MAE & AdaAR & tjAR & AR & MAE-Diff & DiffComp \\
\midrule
ViT3D & 8.43 & 6.29 & 6.45 & 7.66 & 7.16 & 5.84 \\
DisMo & 5.28 & 6.29 & 6.45 & 6.70 & 5.43 & 6.65 \\
TT1D & 10.91 & 9.04 & 8.58 & 7.66 & 13.71 & 10.61 \\
TT3D & 15.38 & 9.04 & 11.42 & 10.30 & \underline{15.99} & \textbf{16.35} \\
\bottomrule
\end{tabular}
\end{adjustbox}
\end{minipage}\hfill
\begin{minipage}[t]{0.485\linewidth}\centering
\caption{End-to-end finetuning of the sweep cells on EPIC-Kitchens-100 verb classification. ViT3D and DisMo were not finetuned.}
\label{tab:app_sweep_ft_ekv}
\setlength{\tabcolsep}{3pt}\renewcommand{\arraystretch}{0.88}
\begin{adjustbox}{max width=\linewidth}
\begin{tabular}{lrrrrrr}
\toprule
Arch & MAE & AdaAR & tjAR & AR & MAE-Diff & DiffComp \\
\midrule
ViT3D & - & - & - & - & - & - \\
DisMo & - & - & - & - & - & - \\
TT1D & 37.92 & 37.82 & 36.07 & 38.54 & 39.24 & 38.23 \\
TT3D & \underline{42.78} & 36.61 & 37.29 & 36.19 & \textbf{43.80} & 42.49 \\
\bottomrule
\end{tabular}
\end{adjustbox}
\end{minipage}
\end{table}

\begin{table}[htbp]\centering\scriptsize
\caption{Architecture-objective sweep on HMDB51 (left) and ARID (right), top-1 accuracy (\%) under all six frozen probes. \textbf{Bold} and \underline{underline} mark the best and second-best cell within each probe and dataset.}
\label{tab:app_sweep_hmdb51_arid}
\setlength{\tabcolsep}{2.2pt}\renewcommand{\arraystretch}{0.88}
\begin{adjustbox}{max width=\linewidth}
\begin{tabular}{llrrrrrr@{\hspace{4pt}}c@{\hspace{4pt}}rrrrrr}
\toprule
 & & \multicolumn{6}{c}{HMDB51} && \multicolumn{6}{c}{ARID} \\
\cmidrule(lr){3-8}\cmidrule(lr){10-15}
Probe & Arch & MAE & AdaAR & tjAR & AR & MAE-Diff & DiffComp && MAE & AdaAR & tjAR & AR & MAE-Diff & DiffComp \\
\midrule
$k$NN & ViT3D & \cellcolor{vmaecolor}\textbf{20.19} & 10.46 & 13.24 & 12.51 & \underline{18.20} & 13.50 && \cellcolor{vmaecolor}\textbf{17.50} & 9.30 & 9.95 & 10.00 & \underline{17.40} & 12.63 \\
 & DisMo & 12.77 & \cellcolor{dismocolor}12.05 & 12.11 & 11.98 & 11.71 & 11.58 && 12.78 & \cellcolor{dismocolor}11.09 & 13.58 & 10.64 & 13.13 & 12.18 \\
 & TT1D & 5.89 & 9.53 & 8.93 & 10.46 & 4.90 & \cellcolor{ourscolor}4.90 && 12.38 & 5.72 & 8.50 & 9.85 & 11.24 & \cellcolor{ourscolor}12.18 \\
 & TT3D & 5.10 & 9.00 & 9.07 & 9.13 & 5.03 & \cellcolor{ourscolor}9.79 && 10.64 & 6.32 & 7.41 & 10.49 & 12.78 & \cellcolor{ourscolor}16.21 \\
\midrule
linear & ViT3D & \cellcolor{vmaecolor}\textbf{29.65} & 17.21 & 19.72 & 18.40 & \underline{27.47} & 19.46 && \cellcolor{vmaecolor}\textbf{21.98} & 11.93 & 11.69 & 10.94 & 17.90 & 14.42 \\
 & DisMo & 20.71 & \cellcolor{dismocolor}18.80 & 18.07 & 17.14 & 19.26 & 17.87 && 17.45 & \cellcolor{dismocolor}12.28 & 15.32 & 11.29 & 16.21 & 16.41 \\
 & TT1D & 11.05 & 13.83 & 11.25 & 12.84 & 10.52 & \cellcolor{ourscolor}10.66 && 14.52 & 6.66 & 12.73 & 6.17 & 11.88 & \cellcolor{ourscolor}15.71 \\
 & TT3D & 11.65 & 12.64 & 11.85 & 13.30 & 11.18 & \cellcolor{ourscolor}12.11 && 14.82 & 10.84 & 11.34 & 5.82 & 10.89 & \cellcolor{ourscolor}\underline{20.69} \\
\midrule
linear$_{\mathrm{lw}}$ & ViT3D & \cellcolor{vmaecolor}\textbf{29.12} & 17.54 & 19.99 & 18.73 & \underline{27.20} & 19.13 && \cellcolor{vmaecolor}\textbf{22.48} & 11.93 & 10.49 & 12.48 & 17.90 & 13.03 \\
 & DisMo & 20.45 & \cellcolor{dismocolor}18.66 & 18.27 & 17.47 & 20.91 & 17.34 && 16.46 & \cellcolor{dismocolor}10.99 & 15.17 & 11.74 & 16.16 & 13.67 \\
 & TT1D & 12.84 & 13.63 & 11.65 & 12.91 & 11.65 & \cellcolor{ourscolor}10.13 && 14.62 & 9.90 & 11.34 & 6.02 & 13.28 & \cellcolor{ourscolor}16.01 \\
 & TT3D & 11.78 & 12.38 & 12.51 & 12.97 & 11.38 & \cellcolor{ourscolor}12.51 && 16.56 & 8.70 & 11.24 & 6.22 & 9.60 & \cellcolor{ourscolor}\underline{20.14} \\
\midrule
MLP & ViT3D & \cellcolor{vmaecolor}\textbf{27.93} & 16.81 & 19.92 & 18.73 & \underline{26.80} & 20.98 && \cellcolor{vmaecolor}\textbf{22.13} & 14.87 & 17.01 & 15.37 & 16.36 & \underline{20.64} \\
 & DisMo & 18.40 & \cellcolor{dismocolor}18.20 & 16.88 & 17.14 & 16.28 & 17.34 && 16.91 & \cellcolor{dismocolor}16.26 & 14.57 & 15.86 & 14.47 & 16.81 \\
 & TT1D & 8.93 & 13.10 & 12.11 & 12.57 & 6.55 & \cellcolor{ourscolor}7.21 && 14.87 & 13.87 & 14.97 & 14.57 & 15.27 & \cellcolor{ourscolor}14.87 \\
 & TT3D & 6.55 & 13.96 & 11.45 & 11.71 & 7.81 & \cellcolor{ourscolor}12.24 && 10.94 & 17.21 & 15.17 & 13.33 & 8.06 & \cellcolor{ourscolor}20.09 \\
\midrule
MLP$_{\mathrm{lw}}$ & ViT3D & \cellcolor{vmaecolor}\textbf{26.74} & 17.87 & 20.05 & 18.13 & \underline{26.41} & 22.44 && \cellcolor{vmaecolor}\textbf{21.63} & 14.52 & 17.75 & 14.12 & 18.25 & \underline{20.59} \\
 & DisMo & 19.39 & \cellcolor{dismocolor}17.14 & 16.68 & 15.35 & 18.40 & 16.81 && 16.16 & \cellcolor{dismocolor}14.87 & 15.61 & 13.97 & 15.71 & 15.56 \\
 & TT1D & 9.33 & 11.98 & 13.50 & 12.57 & 5.96 & \cellcolor{ourscolor}7.48 && 13.87 & 13.72 & 13.13 & 17.75 & 13.92 & \cellcolor{ourscolor}15.42 \\
 & TT3D & 6.88 & 13.10 & 10.92 & 12.31 & 7.15 & \cellcolor{ourscolor}12.57 && 12.03 & 14.72 & 14.77 & 13.58 & 9.85 & \cellcolor{ourscolor}19.34 \\
\midrule
attentive & ViT3D & \cellcolor{vmaecolor}\underline{27.73} & 16.68 & 19.32 & 18.46 & \textbf{27.93} & 20.19 && \cellcolor{vmaecolor}\textbf{24.37} & 13.58 & 15.66 & 14.12 & 18.75 & 17.70 \\
 & DisMo & 17.54 & \cellcolor{dismocolor}17.87 & 17.74 & 16.48 & 18.86 & 15.42 && 18.15 & \cellcolor{dismocolor}15.46 & 14.87 & 14.02 & 15.71 & 16.51 \\
 & TT1D & 10.06 & 13.24 & 13.83 & 13.24 & 6.88 & \cellcolor{ourscolor}8.74 && 17.65 & 13.03 & 12.73 & 13.38 & 14.52 & \cellcolor{ourscolor}16.41 \\
 & TT3D & 7.61 & 13.50 & 13.10 & 12.84 & 6.68 & \cellcolor{ourscolor}12.64 && 11.49 & 17.65 & 13.92 & 15.76 & 7.71 & \cellcolor{ourscolor}\underline{21.98} \\
\bottomrule
\end{tabular}
\end{adjustbox}
\end{table}

\begin{table}[htbp]\centering\scriptsize
\caption{Architecture-objective sweep on IARD (left) and Jester (right), top-1 accuracy (\%) under all six frozen probes. \textbf{Bold} and \underline{underline} mark the best and second-best cell within each probe and dataset.}
\label{tab:app_sweep_iard_jester}
\setlength{\tabcolsep}{2.2pt}\renewcommand{\arraystretch}{0.88}
\begin{adjustbox}{max width=\linewidth}
\begin{tabular}{llrrrrrr@{\hspace{4pt}}c@{\hspace{4pt}}rrrrrr}
\toprule
 & & \multicolumn{6}{c}{IARD} && \multicolumn{6}{c}{Jester} \\
\cmidrule(lr){3-8}\cmidrule(lr){10-15}
Probe & Arch & MAE & AdaAR & tjAR & AR & MAE-Diff & DiffComp && MAE & AdaAR & tjAR & AR & MAE-Diff & DiffComp \\
\midrule
$k$NN & ViT3D & \cellcolor{vmaecolor}63.96 & 40.99 & 61.21 & 41.98 & 64.73 & \underline{71.87} && \cellcolor{vmaecolor}14.30 & 6.38 & 5.55 & 5.92 & \underline{14.59} & 11.27 \\
 & DisMo & 50.55 & \cellcolor{dismocolor}57.25 & 49.45 & 46.81 & 57.36 & 50.55 && 9.04 & \cellcolor{dismocolor}8.68 & 9.76 & 6.21 & 9.37 & 8.02 \\
 & TT1D & 48.35 & 28.68 & 24.62 & 27.80 & 41.54 & \cellcolor{ourscolor}49.23 && 11.21 & 4.63 & 4.34 & 4.67 & 8.64 & \cellcolor{ourscolor}10.25 \\
 & TT3D & 35.16 & 34.62 & 33.19 & 26.04 & 28.46 & \cellcolor{ourscolor}\textbf{76.70} && 5.69 & 5.46 & 4.47 & 4.60 & 5.26 & \cellcolor{ourscolor}\textbf{18.21} \\
\midrule
linear & ViT3D & \cellcolor{vmaecolor}\textbf{72.31} & 44.51 & 55.27 & 54.73 & \underline{69.12} & 59.12 && \cellcolor{vmaecolor}22.61 & 12.45 & 11.63 & 12.36 & \underline{22.71} & 13.34 \\
 & DisMo & 61.76 & \cellcolor{dismocolor}57.47 & 54.29 & 52.31 & 62.64 & 52.42 && 15.18 & \cellcolor{dismocolor}12.72 & 13.21 & 12.36 & 15.54 & 12.68 \\
 & TT1D & 49.12 & 34.29 & 33.19 & 42.20 & 57.36 & \cellcolor{ourscolor}50.44 && 14.36 & 10.42 & 10.42 & 10.42 & 12.22 & \cellcolor{ourscolor}13.93 \\
 & TT3D & 52.75 & 32.53 & 49.23 & 36.26 & 28.68 & \cellcolor{ourscolor}\underline{69.12} && 12.19 & 10.42 & 10.42 & 10.42 & 11.76 & \cellcolor{ourscolor}\textbf{26.91} \\
\midrule
linear$_{\mathrm{lw}}$ & ViT3D & \cellcolor{vmaecolor}\textbf{71.10} & 42.75 & 51.43 & 54.29 & \underline{70.22} & 60.11 && \cellcolor{vmaecolor}22.67 & 12.45 & 11.63 & 12.22 & \underline{23.73} & 13.31 \\
 & DisMo & 62.09 & \cellcolor{dismocolor}57.80 & 55.93 & 52.31 & 65.60 & 56.26 && 15.64 & \cellcolor{dismocolor}12.85 & 13.08 & 12.39 & 17.75 & 12.65 \\
 & TT1D & 49.12 & 33.08 & 37.14 & 42.53 & 52.20 & \cellcolor{ourscolor}50.33 && 15.74 & 10.42 & 10.42 & 10.42 & 14.53 & \cellcolor{ourscolor}15.15 \\
 & TT3D & 52.75 & 31.32 & 46.59 & 42.09 & 24.40 & \cellcolor{ourscolor}69.56 && 13.01 & 10.42 & 10.42 & 10.42 & 13.21 & \cellcolor{ourscolor}\textbf{29.54} \\
\midrule
MLP & ViT3D & \cellcolor{vmaecolor}\textbf{74.73} & 57.47 & 69.67 & 61.21 & \underline{70.88} & 66.48 && \cellcolor{vmaecolor}\underline{31.94} & 13.08 & 14.20 & 12.65 & 30.00 & 18.80 \\
 & DisMo & 57.91 & \cellcolor{dismocolor}56.37 & 57.47 & 51.10 & 58.57 & 61.98 && 20.08 & \cellcolor{dismocolor}15.12 & 15.51 & 13.54 & 19.36 & 12.72 \\
 & TT1D & 56.15 & 53.63 & 40.33 & 60.55 & 58.90 & \cellcolor{ourscolor}60.44 && 18.24 & 10.45 & 10.42 & 10.45 & 13.70 & \cellcolor{ourscolor}16.37 \\
 & TT3D & 33.85 & 44.84 & 46.04 & 53.41 & 28.68 & \cellcolor{ourscolor}67.69 && 11.40 & 10.42 & 10.42 & 10.45 & 9.83 & \cellcolor{ourscolor}\textbf{32.86} \\
\midrule
MLP$_{\mathrm{lw}}$ & ViT3D & \cellcolor{vmaecolor}\textbf{76.48} & 56.59 & \underline{70.44} & 63.74 & 70.33 & 68.79 && \cellcolor{vmaecolor}\underline{32.04} & 13.05 & 14.23 & 12.82 & 30.66 & 18.40 \\
 & DisMo & 57.80 & \cellcolor{dismocolor}58.02 & 58.57 & 51.87 & 59.67 & 61.21 && 20.37 & \cellcolor{dismocolor}15.25 & 15.77 & 13.77 & 18.11 & 12.45 \\
 & TT1D & 57.03 & 53.08 & 43.19 & 60.77 & 59.56 & \cellcolor{ourscolor}60.22 && 18.63 & 10.42 & 10.42 & 10.48 & 15.28 & \cellcolor{ourscolor}16.79 \\
 & TT3D & 37.36 & 46.81 & 45.05 & 54.18 & 32.86 & \cellcolor{ourscolor}68.35 && 11.73 & 10.78 & 10.45 & 10.45 & 10.65 & \cellcolor{ourscolor}\textbf{35.49} \\
\midrule
attentive & ViT3D & \cellcolor{vmaecolor}\textbf{80.44} & 53.41 & 62.86 & 54.95 & 74.95 & \underline{77.03} && \cellcolor{vmaecolor}\underline{39.47} & 12.85 & 13.01 & 12.85 & 31.68 & 19.82 \\
 & DisMo & 73.19 & \cellcolor{dismocolor}56.81 & 58.79 & 57.25 & 54.84 & 59.12 && 29.71 & \cellcolor{dismocolor}16.46 & 15.94 & 13.54 & 21.66 & 12.98 \\
 & TT1D & 62.42 & 38.46 & 40.00 & 28.35 & 57.14 & \cellcolor{ourscolor}76.48 && 26.65 & 11.70 & 10.45 & 11.86 & 15.68 & \cellcolor{ourscolor}24.32 \\
 & TT3D & 24.07 & 39.45 & 24.07 & 23.96 & 32.09 & \cellcolor{ourscolor}67.03 && 11.07 & 12.85 & 10.42 & 11.83 & 10.32 & \cellcolor{ourscolor}\textbf{53.89} \\
\bottomrule
\end{tabular}
\end{adjustbox}
\end{table}

\begin{table}[htbp]\centering\scriptsize
\caption{Architecture-objective sweep on Something-Something\,V2 (left) and Diving48 (frozen) (right), top-1 accuracy (\%) under all six frozen probes. \textbf{Bold} and \underline{underline} mark the best and second-best cell within each probe and dataset.}
\label{tab:app_sweep_sthsthv2_diving48}
\setlength{\tabcolsep}{2.2pt}\renewcommand{\arraystretch}{0.88}
\begin{adjustbox}{max width=\linewidth}
\begin{tabular}{llrrrrrr@{\hspace{4pt}}c@{\hspace{4pt}}rrrrrr}
\toprule
 & & \multicolumn{6}{c}{Something-Something\,V2} && \multicolumn{6}{c}{Diving48 (frozen)} \\
\cmidrule(lr){3-8}\cmidrule(lr){10-15}
Probe & Arch & MAE & AdaAR & tjAR & AR & MAE-Diff & DiffComp && MAE & AdaAR & tjAR & AR & MAE-Diff & DiffComp \\
\midrule
$k$NN & ViT3D & \cellcolor{vmaecolor}\textbf{4.83} & 1.44 & 1.61 & 1.51 & 4.24 & 1.99 && \cellcolor{vmaecolor}\underline{8.58} & 7.41 & 6.95 & 8.32 & \textbf{8.98} & 6.95 \\
 & DisMo & 1.90 & \cellcolor{dismocolor}1.67 & 1.69 & 1.42 & 1.89 & 1.67 && 6.60 & \cellcolor{dismocolor}7.66 & 8.07 & 7.21 & 6.45 & 7.92 \\
 & TT1D & 2.12 & 1.12 & 1.10 & 1.07 & 1.67 & \cellcolor{ourscolor}2.30 && 6.09 & 6.35 & 6.40 & 6.04 & 5.08 & \cellcolor{ourscolor}5.23 \\
 & TT3D & 1.26 & 1.08 & 1.10 & 1.14 & 1.33 & \cellcolor{ourscolor}\underline{4.63} && 4.92 & 6.65 & 6.75 & 6.45 & 5.28 & \cellcolor{ourscolor}6.14 \\
\midrule
linear & ViT3D & \cellcolor{vmaecolor}\textbf{9.94} & 2.66 & 2.99 & 2.73 & 9.41 & 4.35 && \cellcolor{vmaecolor}8.22 & 7.82 & 7.36 & 8.38 & 7.21 & \textbf{9.09} \\
 & DisMo & 5.64 & \cellcolor{dismocolor}2.95 & 3.48 & 2.39 & 6.22 & 2.93 && 7.46 & \cellcolor{dismocolor}7.41 & 7.72 & 7.31 & 8.07 & \underline{8.73} \\
 & TT1D & 5.16 & 1.93 & 1.60 & 1.79 & 4.06 & \cellcolor{ourscolor}4.56 && 6.55 & 6.40 & 5.63 & 5.89 & 5.94 & \cellcolor{ourscolor}6.75 \\
 & TT3D & 3.29 & 1.74 & 1.82 & 1.78 & 3.45 & \cellcolor{ourscolor}\underline{9.80} && 6.60 & 5.43 & 5.94 & 5.48 & 6.80 & \cellcolor{ourscolor}7.77 \\
\midrule
linear$_{\mathrm{lw}}$ & ViT3D & \cellcolor{vmaecolor}\textbf{10.41} & 2.72 & 3.02 & 2.58 & 9.44 & 4.29 && \cellcolor{vmaecolor}8.48 & 7.01 & \underline{8.88} & 7.77 & \textbf{9.29} & 8.22 \\
 & DisMo & 6.06 & \cellcolor{dismocolor}3.25 & 3.64 & 2.59 & 6.48 & 3.02 && 8.43 & \cellcolor{dismocolor}7.41 & 8.07 & 7.31 & 8.78 & 7.26 \\
 & TT1D & 5.50 & 1.98 & 1.89 & 1.94 & 4.46 & \cellcolor{ourscolor}5.32 && 6.14 & 6.24 & 5.63 & 6.04 & 6.40 & \cellcolor{ourscolor}5.69 \\
 & TT3D & 3.89 & 1.73 & 1.77 & 1.81 & 3.79 & \cellcolor{ourscolor}\underline{10.28} && 6.95 & 5.94 & 6.04 & 6.19 & 6.04 & \cellcolor{ourscolor}6.60 \\
\midrule
MLP & ViT3D & \cellcolor{vmaecolor}\textbf{12.59} & 4.23 & 4.34 & 3.66 & \underline{11.51} & 5.88 && \cellcolor{vmaecolor}\underline{9.14} & 8.73 & 8.83 & 7.97 & 8.93 & 7.26 \\
 & DisMo & 7.05 & \cellcolor{dismocolor}4.16 & 4.15 & 3.31 & 7.94 & 3.59 && 7.72 & \cellcolor{dismocolor}7.87 & 8.43 & \textbf{9.19} & 8.73 & 8.43 \\
 & TT1D & 5.34 & 2.21 & 1.95 & 2.40 & 5.12 & \cellcolor{ourscolor}4.75 && 5.48 & 6.95 & 5.89 & 5.94 & 4.87 & \cellcolor{ourscolor}5.84 \\
 & TT3D & 4.21 & 2.18 & 2.04 & 2.12 & 4.44 & \cellcolor{ourscolor}11.39 && 5.48 & 6.45 & 7.56 & 6.70 & 5.74 & \cellcolor{ourscolor}5.63 \\
\midrule
MLP$_{\mathrm{lw}}$ & ViT3D & \cellcolor{vmaecolor}\textbf{12.54} & 4.14 & 4.55 & 3.73 & \underline{11.85} & 5.92 && \cellcolor{vmaecolor}\textbf{9.19} & 7.56 & 7.92 & 8.48 & 8.43 & 8.48 \\
 & DisMo & 7.29 & \cellcolor{dismocolor}4.33 & 4.97 & 3.25 & 8.27 & 3.80 && \underline{8.93} & \cellcolor{dismocolor}7.66 & 8.17 & 8.32 & 8.53 & 7.41 \\
 & TT1D & 5.17 & 2.14 & 2.28 & 2.41 & 5.78 & \cellcolor{ourscolor}5.17 && 5.53 & 6.55 & 6.75 & 6.35 & 4.72 & \cellcolor{ourscolor}6.40 \\
 & TT3D & 4.62 & 2.40 & 2.04 & 2.23 & 4.63 & \cellcolor{ourscolor}11.69 && 5.84 & 6.14 & 7.72 & 6.45 & 6.65 & \cellcolor{ourscolor}6.35 \\
\midrule
attentive & ViT3D & \cellcolor{vmaecolor}\underline{14.95} & 3.99 & 4.86 & 3.93 & 11.73 & 6.57 && \cellcolor{vmaecolor}8.68 & 7.11 & 7.87 & 8.83 & \underline{8.93} & 7.16 \\
 & DisMo & 11.20 & \cellcolor{dismocolor}5.29 & 4.35 & 3.54 & 7.89 & 5.53 && 8.17 & \cellcolor{dismocolor}\textbf{9.19} & 8.17 & 8.58 & 8.32 & 6.70 \\
 & TT1D & 10.05 & 4.71 & 2.53 & 4.86 & 6.00 & \cellcolor{ourscolor}7.91 && 5.69 & 7.06 & 7.06 & 6.14 & 5.18 & \cellcolor{ourscolor}5.58 \\
 & TT3D & 4.98 & 2.26 & 4.16 & 4.39 & 4.82 & \cellcolor{ourscolor}\textbf{18.28} && 6.90 & 7.26 & 8.88 & 6.29 & 6.45 & \cellcolor{ourscolor}5.63 \\
\bottomrule
\end{tabular}
\end{adjustbox}
\end{table}

\begin{table}[htbp]\centering\scriptsize
\caption{Architecture-objective sweep on EK-100 verb (left) and EK-100 verb anticipation (right), top-1 accuracy (\%) under all six frozen probes. \textbf{Bold} and \underline{underline} mark the best and second-best cell within each probe and dataset. TT1D was not evaluated on anticipation.}
\label{tab:app_sweep_ekv_ekv_antic}
\setlength{\tabcolsep}{2.2pt}\renewcommand{\arraystretch}{0.88}
\begin{adjustbox}{max width=\linewidth}
\begin{tabular}{llrrrrrr@{\hspace{4pt}}c@{\hspace{4pt}}rrrrrr}
\toprule
 & & \multicolumn{6}{c}{EK-100 verb} && \multicolumn{6}{c}{EK-100 verb anticipation} \\
\cmidrule(lr){3-8}\cmidrule(lr){10-15}
Probe & Arch & MAE & AdaAR & tjAR & AR & MAE-Diff & DiffComp && MAE & AdaAR & tjAR & AR & MAE-Diff & DiffComp \\
\midrule
$k$NN & ViT3D & \cellcolor{vmaecolor}\textbf{31.01} & 27.32 & 29.41 & 27.81 & \underline{30.79} & 28.18 && \cellcolor{vmaecolor}20.68 & 20.29 & \textbf{21.83} & 20.18 & \underline{21.33} & 20.15 \\
 & DisMo & 26.04 & \cellcolor{dismocolor}25.52 & 25.40 & 25.46 & 25.71 & 25.31 && 20.52 & \cellcolor{dismocolor}19.85 & 20.17 & 20.29 & 20.97 & 19.94 \\
 & TT1D & 20.12 & 21.75 & 21.35 & 22.60 & 19.78 & \cellcolor{ourscolor}19.45 && - & - & - & - & - & \cellcolor{ourscolor}- \\
 & TT3D & 19.97 & 21.39 & 21.48 & 21.57 & 20.35 & \cellcolor{ourscolor}21.97 && 19.40 & 19.10 & 18.76 & 18.93 & 19.56 & \cellcolor{ourscolor}19.85 \\
\midrule
linear & ViT3D & \cellcolor{vmaecolor}\textbf{31.77} & 23.20 & 23.75 & 22.36 & \underline{30.67} & 26.85 && \cellcolor{vmaecolor}22.09 & 21.81 & 22.61 & 22.13 & \textbf{23.32} & 22.62 \\
 & DisMo & 25.34 & \cellcolor{dismocolor}21.64 & 22.70 & 21.87 & 26.94 & 22.70 && 21.15 & \cellcolor{dismocolor}\underline{22.68} & 21.70 & 22.62 & 21.49 & 20.76 \\
 & TT1D & 23.32 & 20.78 & 19.68 & 20.21 & 21.71 & \cellcolor{ourscolor}21.35 && - & - & - & - & - & \cellcolor{ourscolor}- \\
 & TT3D & 21.34 & 19.66 & 19.83 & 20.14 & 21.43 & \cellcolor{ourscolor}25.22 && 21.19 & 20.47 & 20.28 & 20.42 & 21.67 & \cellcolor{ourscolor}20.64 \\
\midrule
linear$_{\mathrm{lw}}$ & ViT3D & \cellcolor{vmaecolor}\textbf{31.84} & 22.97 & 25.42 & 22.37 & \underline{31.08} & 26.48 && \cellcolor{vmaecolor}22.96 & \underline{23.07} & 22.42 & 22.23 & \textbf{23.22} & 22.76 \\
 & DisMo & 27.85 & \cellcolor{dismocolor}23.20 & 22.68 & 21.57 & 26.14 & 23.12 && 22.25 & \cellcolor{dismocolor}22.23 & 21.67 & 22.66 & 21.89 & 21.39 \\
 & TT1D & 23.75 & 20.71 & 19.86 & 20.10 & 21.59 & \cellcolor{ourscolor}21.18 && - & - & - & - & - & \cellcolor{ourscolor}- \\
 & TT3D & 21.66 & 19.82 & 19.87 & 20.12 & 21.91 & \cellcolor{ourscolor}25.81 && 21.57 & 20.49 & 20.64 & 20.73 & 21.20 & \cellcolor{ourscolor}20.92 \\
\midrule
MLP & ViT3D & \cellcolor{vmaecolor}\textbf{34.99} & 25.82 & 26.32 & 25.22 & \underline{33.69} & 29.11 && \cellcolor{vmaecolor}22.66 & 22.92 & 22.59 & 22.46 & \textbf{23.62} & 23.07 \\
 & DisMo & 27.80 & \cellcolor{dismocolor}25.39 & 24.43 & 23.68 & 29.78 & 24.40 && 22.34 & \cellcolor{dismocolor}\underline{23.13} & 22.26 & 22.53 & 22.05 & 22.55 \\
 & TT1D & 23.42 & 22.36 & 20.11 & 22.35 & 22.20 & \cellcolor{ourscolor}21.46 && - & - & - & - & - & \cellcolor{ourscolor}- \\
 & TT3D & 21.42 & 20.30 & 20.51 & 22.30 & 22.60 & \cellcolor{ourscolor}27.13 && 20.93 & 20.96 & 20.93 & 21.12 & 21.52 & \cellcolor{ourscolor}20.08 \\
\midrule
MLP$_{\mathrm{lw}}$ & ViT3D & \cellcolor{vmaecolor}\textbf{34.54} & 25.85 & 27.60 & 26.16 & \underline{33.84} & 28.43 && \cellcolor{vmaecolor}\textbf{23.31} & 22.53 & 22.73 & 22.59 & \underline{23.27} & 23.03 \\
 & DisMo & 28.27 & \cellcolor{dismocolor}24.58 & 24.91 & 23.51 & 27.55 & 23.87 && 21.57 & \cellcolor{dismocolor}22.34 & 22.22 & 22.59 & 22.11 & 22.93 \\
 & TT1D & 24.11 & 22.08 & 20.17 & 22.16 & 24.04 & \cellcolor{ourscolor}21.86 && - & - & - & - & - & \cellcolor{ourscolor}- \\
 & TT3D & 22.28 & 22.24 & 21.39 & 21.99 & 22.31 & \cellcolor{ourscolor}27.20 && 21.12 & 21.36 & 20.61 & 21.23 & 20.49 & \cellcolor{ourscolor}19.40 \\
\midrule
attentive & ViT3D & \cellcolor{vmaecolor}\textbf{35.62} & 26.43 & 27.58 & 25.48 & \underline{33.62} & 29.24 && \cellcolor{vmaecolor}22.81 & \underline{23.47} & 22.97 & 22.62 & \textbf{23.53} & 23.39 \\
 & DisMo & 27.77 & \cellcolor{dismocolor}25.20 & 25.48 & 23.41 & 29.79 & 25.33 && 22.51 & \cellcolor{dismocolor}23.30 & 22.86 & 22.81 & 21.99 & 22.47 \\
 & TT1D & 24.28 & 22.79 & 21.34 & 22.87 & 24.46 & \cellcolor{ourscolor}22.75 && - & - & - & - & - & \cellcolor{ourscolor}- \\
 & TT3D & 22.72 & 21.85 & 21.28 & 22.07 & 24.19 & \cellcolor{ourscolor}28.23 && 20.30 & 20.85 & 20.94 & 21.12 & 21.13 & \cellcolor{ourscolor}19.34 \\
\bottomrule
\end{tabular}
\end{adjustbox}
\end{table}

\subsection{Decoder Ablation under Every Probe}
\label{sec:appendix-complete-decoder}

\begin{figure}[htbp]\centering
\includegraphics[width=\textwidth]{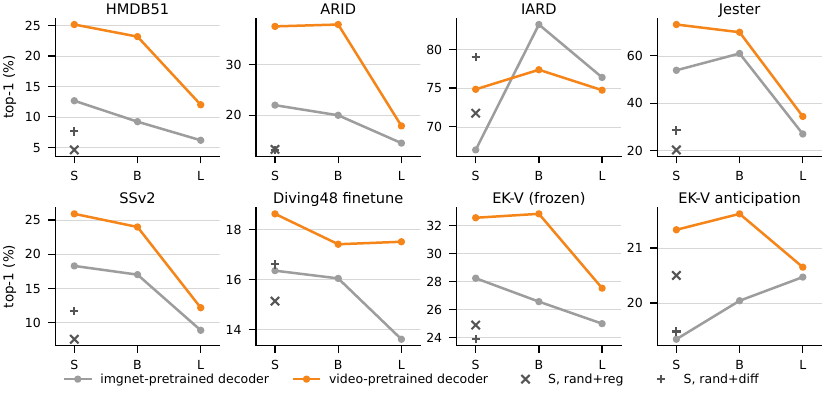}
\caption{Decoder size and initialization on TT3D + Diff Compression, attentive probe (finetune for Diving48). Crosses are size-S decoders from random initialization.}
\label{fig:app_decoder_scaling}
\end{figure}

Figure~\ref{fig:app_decoder_scaling} plots the decoder ablation of Section~\ref{sec:exp-decoder} on every task. The video-pretrained decoder is above the ImageNet-pretrained one at every size on the motion-heavy tasks, and size S or B gives the best result on every task. Table~\ref{tab:app_decoder_probes} repeats the ablation under every probe, Table~\ref{tab:app_decoder_downstream} on the downstream tasks, and Table~\ref{tab:app_decoder_init} separates decoder pretraining from the objective.

\begin{table}[htbp]\centering\scriptsize
\caption{Decoder ablation on TT3D + Diff Compression under all six frozen probes. Rows name the decoder size and its initialization (rand+reg and rand+diff are size S from random initialization with a regression or diffusion auxiliary loss). Best and second-best per probe and column marked.}
\label{tab:app_decoder_probes}
\begin{minipage}[t]{0.495\linewidth}\centering
\setlength{\tabcolsep}{3pt}\renewcommand{\arraystretch}{0.88}
\begin{adjustbox}{max width=\linewidth}
\begin{tabular}{llrrrrr}
\toprule
Probe & Decoder & HMDB51 & ARID & IARD & Jester & SSv2 \\
\midrule
$k$NN & rand+reg & 3.90 & 9.60 & 56.48 & 6.93 & 1.46 \\
 & rand+diff & 9.00 & 12.58 & \underline{74.84} & 17.42 & 3.93 \\
 & S imgnet & 9.79 & 16.21 & \textbf{76.70} & 18.21 & 4.63 \\
 & S video & \textbf{13.57} & \textbf{21.81} & 62.20 & \textbf{27.05} & \textbf{9.11} \\
 & B imgnet & 8.47 & 14.54 & 71.87 & 15.41 & 2.60 \\
 & B video & \underline{13.30} & \underline{20.04} & 66.26 & \underline{22.57} & \underline{6.89} \\
 & L imgnet & 4.96 & 12.38 & 73.30 & 11.76 & 1.95 \\
 & L video & 8.93 & 13.63 & 71.10 & 14.69 & 2.58 \\
\midrule
linear & rand+reg & 7.94 & 14.12 & 70.66 & 17.55 & 5.73 \\
 & rand+diff & 10.52 & 13.92 & 71.98 & 23.89 & 8.11 \\
 & S imgnet & 12.11 & 20.69 & 69.12 & 26.91 & 9.80 \\
 & S video & \underline{23.23} & \underline{27.69} & 64.73 & \textbf{38.82} & \textbf{14.44} \\
 & B imgnet & 10.46 & 15.77 & \underline{76.04} & 28.14 & 7.83 \\
 & B video & \textbf{24.88} & \textbf{28.43} & 70.44 & \underline{35.36} & \underline{13.19} \\
 & L imgnet & 10.52 & 16.76 & 72.42 & 21.26 & 6.52 \\
 & L video & 16.02 & 18.50 & \textbf{78.46} & 26.32 & 7.60 \\
\midrule
linear$_{\mathrm{lw}}$ & rand+reg & 8.01 & 13.67 & 74.18 & 18.14 & 6.43 \\
 & rand+diff & 12.97 & 12.13 & 72.64 & 25.57 & 8.48 \\
 & S imgnet & 12.51 & 20.14 & 69.56 & 29.54 & 10.28 \\
 & S video & \underline{23.10} & \underline{27.79} & 65.38 & \textbf{42.73} & \textbf{15.22} \\
 & B imgnet & 10.59 & 16.52 & \underline{77.14} & 31.20 & 8.31 \\
 & B video & \textbf{24.95} & \textbf{29.34} & 69.89 & \underline{39.05} & \underline{13.86} \\
 & L imgnet & 12.24 & 16.81 & 73.63 & 22.38 & 6.78 \\
 & L video & 15.62 & 17.90 & \textbf{79.23} & 28.62 & 8.16 \\
\bottomrule
\end{tabular}
\end{adjustbox}
\end{minipage}\hfill
\begin{minipage}[t]{0.495\linewidth}\centering
\setlength{\tabcolsep}{3pt}\renewcommand{\arraystretch}{0.88}
\begin{adjustbox}{max width=\linewidth}
\begin{tabular}{llrrrrr}
\toprule
Probe & Decoder & HMDB51 & ARID & IARD & Jester & SSv2 \\
\midrule
MLP & rand+reg & 5.43 & 13.97 & 69.56 & 20.74 & 7.31 \\
 & rand+diff & 7.68 & 13.13 & 76.48 & 26.91 & 10.53 \\
 & S imgnet & 12.24 & 20.09 & 67.69 & 32.86 & 11.39 \\
 & S video & \underline{21.24} & \underline{30.89} & 67.80 & \textbf{49.39} & \textbf{18.68} \\
 & B imgnet & 9.46 & 17.48 & \textbf{78.02} & 37.55 & 10.72 \\
 & B video & \textbf{21.97} & \textbf{31.21} & 75.93 & \underline{46.37} & \underline{17.26} \\
 & L imgnet & 6.35 & 14.37 & \underline{77.58} & 24.91 & 8.62 \\
 & L video & 11.25 & 17.65 & 73.96 & 31.02 & 11.39 \\
\midrule
MLP$_{\mathrm{lw}}$ & rand+reg & 5.03 & 13.23 & 71.32 & 20.93 & 7.79 \\
 & rand+diff & 7.74 & 13.53 & \textbf{79.23} & 27.90 & 11.46 \\
 & S imgnet & 12.57 & 19.34 & 68.35 & 35.49 & 11.69 \\
 & S video & \underline{20.98} & \underline{29.45} & 67.58 & \textbf{53.39} & \textbf{19.33} \\
 & B imgnet & 9.33 & 16.68 & \underline{76.92} & 41.35 & 11.37 \\
 & B video & \textbf{22.57} & \textbf{31.43} & \underline{76.92} & \underline{50.90} & \underline{18.29} \\
 & L imgnet & 6.29 & 13.03 & 76.15 & 27.05 & 8.55 \\
 & L video & 9.73 & 18.10 & 72.97 & 34.11 & 11.51 \\
\midrule
attentive & rand+reg & 4.57 & 13.28 & 71.76 & 20.21 & 7.53 \\
 & rand+diff & 7.61 & 13.18 & \underline{79.01} & 28.46 & 11.71 \\
 & S imgnet & 12.64 & 21.98 & 67.03 & 53.89 & 18.28 \\
 & S video & \textbf{25.15} & \underline{37.47} & 74.84 & \textbf{73.25} & \textbf{25.92} \\
 & B imgnet & 9.20 & 19.99 & \textbf{83.19} & 61.00 & 17.00 \\
 & B video & \underline{23.16} & \textbf{37.84} & 77.36 & \underline{69.99} & \underline{23.99} \\
 & L imgnet & 6.15 & 14.52 & 76.37 & 26.95 & 8.84 \\
 & L video & 11.98 & 17.90 & 74.73 & 34.37 & 12.15 \\
\bottomrule
\end{tabular}
\end{adjustbox}
\end{minipage}
\end{table}

\begin{table}[htbp]\scriptsize
\begin{minipage}[t]{0.39\linewidth}\centering
\caption{Decoder ablation on the downstream tasks: frozen attentive probe and end-to-end finetune (FT) on Diving48 and EPIC-Kitchens-100 verb classification, and frozen verb anticipation.}
\label{tab:app_decoder_downstream}
\setlength{\tabcolsep}{3pt}\renewcommand{\arraystretch}{0.88}
\begin{adjustbox}{max width=\linewidth}
\begin{tabular}{lrrrrr}
\toprule
 & \multicolumn{2}{c}{Diving48} & \multicolumn{2}{c}{EK-V} & Antic. \\
\cmidrule(lr){2-3}\cmidrule(lr){4-5}\cmidrule(lr){6-6}
Decoder & frozen & FT & frozen & FT & frozen \\
\midrule
rand+reg & 5.53 & 15.13 & 24.89 & 41.68 & 20.50 \\
rand+diff & 6.70 & 16.60 & 23.88 & \textbf{43.18} & 19.48 \\
S imgnet & 5.63 & 16.35 & 28.23 & \underline{42.49} & 19.34 \\
S video & \underline{8.17} & \textbf{18.63} & \underline{32.54} & 40.68 & \underline{21.33} \\
B imgnet & 6.09 & 16.04 & 26.56 & 41.76 & 20.04 \\
B video & \textbf{9.64} & 17.41 & \textbf{32.82} & 42.31 & \textbf{21.62} \\
L imgnet & 5.69 & 13.60 & 24.99 & 40.60 & 20.47 \\
L video & 7.16 & \underline{17.51} & 27.52 & 39.42 & 20.65 \\
\bottomrule
\end{tabular}
\end{adjustbox}
\end{minipage}\hfill
\begin{minipage}[t]{0.585\linewidth}\centering
\caption{TT3D decoder initialization for two objectives: the size-S decoder pretrained on ImageNet against random initialization with a diffusion or regression loss.}
\label{tab:app_decoder_init}
\setlength{\tabcolsep}{3pt}\renewcommand{\arraystretch}{0.88}
\begin{adjustbox}{max width=\linewidth}
\begin{tabular}{lllrrrrr}
\toprule
Probe & Objective & Decoder init & HMDB51 & ARID & IARD & Jester & SSv2 \\
\midrule
$k$NN & DiffComp & pretrained & 9.79 & \textbf{16.21} & \textbf{76.70} & \textbf{18.21} & \textbf{4.63} \\
 & DiffComp & rand+diff & 9.00 & \underline{12.58} & \underline{74.84} & \underline{17.42} & \underline{3.93} \\
 & DiffComp & rand+reg & 3.90 & 9.60 & 56.48 & 6.93 & 1.46 \\
 & AdaAR & pretrained & 9.00 & 6.32 & 34.62 & 5.46 & 1.08 \\
 & AdaAR & rand+diff & \textbf{11.18} & 9.60 & 33.63 & 4.14 & 1.31 \\
 & AdaAR & rand+reg & \underline{10.92} & 8.35 & 50.44 & 6.47 & 1.90 \\
\midrule
attentive & DiffComp & pretrained & 12.64 & \textbf{21.98} & 67.03 & \textbf{53.89} & \textbf{18.28} \\
 & DiffComp & rand+diff & 7.61 & 13.18 & \textbf{79.01} & \underline{28.46} & \underline{11.71} \\
 & DiffComp & rand+reg & 4.57 & 13.28 & \underline{71.76} & 20.21 & 7.53 \\
 & AdaAR & pretrained & 13.50 & \underline{17.65} & 39.45 & 12.85 & 2.26 \\
 & AdaAR & rand+diff & \underline{17.41} & 12.33 & 34.07 & 9.76 & 3.29 \\
 & AdaAR & rand+reg & \textbf{18.20} & 10.69 & 57.47 & 11.07 & 5.63 \\
\bottomrule
\end{tabular}
\end{adjustbox}
\end{minipage}
\end{table}

\subsection{Training-Recipe Variants}
\label{sec:appendix-complete-variants}

Three variants change one element of the recipe. Table~\ref{tab:app_dual_aug} applies DisMo-style dual augmentation to every architecture. It raises Jester for every architecture and DisMo the most, which is why the final comparison uses DisMo in this setting. Table~\ref{tab:app_dismo_decoder} varies the DisMo decoder, and Table~\ref{tab:app_k710} adds Kinetics-710 \citep{openmmlab2024kinetics710} to the pretraining mixture.

\begin{table}[htbp]\centering\scriptsize
\caption{DisMo-style dual augmentation (encoder and decoder see different spatial augmentations) applied to every architecture, size-S ImageNet decoder. Each cell is no augmentation / dual augmentation. (a) and (b) are the two TT3D runs of Table~\ref{tab:app_all_runs}.}
\label{tab:app_dual_aug}
\setlength{\tabcolsep}{3pt}\renewcommand{\arraystretch}{0.88}
\begin{adjustbox}{max width=\linewidth}
\begin{tabular}{lllrrrrr}
\toprule
Probe & Arch & Objective & HMDB51 & ARID & IARD & Jester & SSv2 \\
\midrule
$k$NN & ViT3D & AdaAR & 10.46 / 16.02 & 9.30 / 9.83 & 40.99 / 46.59 & 6.38 / 11.14 & 1.44 / 2.51 \\
 & ViT3D & tjAR & 13.24 / 15.95 & 9.95 / 14.38 & 61.21 / 48.68 & 5.55 / 10.90 & 1.61 / 2.82 \\
 & DisMo & AdaAR & 12.05 / 16.74 & 11.09 / 12.56 & 57.25 / 68.68 & 8.68 / 12.27 & 1.67 / 2.91 \\
 & DisMo & tjAR & 12.11 / 16.41 & 13.58 / 11.81 & 49.45 / 55.82 & 9.76 / 12.58 & 1.69 / 2.70 \\
 & TT1D & AdaAR & 9.53 / 6.75 & 5.72 / 14.22 & 28.68 / 24.40 & 4.63 / 5.53 & 1.12 / 1.45 \\
 & TT1D & tjAR & 8.93 / 6.95 & 8.50 / 7.75 & 24.62 / 27.36 & 4.34 / 4.93 & 1.10 / 1.10 \\
 & TT3D & AdaAR (a) & 9.00 / 12.51 & 6.32 / 10.96 & 34.62 / 53.63 & 5.46 / 5.59 & 1.08 / 1.44 \\
 & TT3D & AdaAR (b) & 9.00 / 11.25 & 6.32 / 11.28 & 34.62 / 45.71 & 5.46 / 5.52 & 1.08 / 1.48 \\
 & TT3D & tjAR & 9.07 / 11.78 & 7.41 / 9.67 & 33.19 / 36.15 & 4.47 / 5.73 & 1.10 / 1.31 \\
\midrule
attentive & ViT3D & AdaAR & 16.68 / 22.96 & 13.58 / 17.64 & 53.41 / 70.77 & 12.85 / 32.62 & 3.99 / 9.78 \\
 & ViT3D & tjAR & 19.32 / 20.78 & 15.66 / 17.85 & 62.86 / 72.86 & 13.01 / 30.61 & 4.86 / 10.05 \\
 & DisMo & AdaAR & 17.87 / 22.57 & 15.46 / 22.45 & 56.81 / 89.89 & 16.46 / 46.95 & 5.29 / 13.33 \\
 & DisMo & tjAR & 17.74 / 24.49 & 14.87 / 21.49 & 58.79 / 85.60 & 15.94 / 48.23 & 4.35 / 13.10 \\
 & TT1D & AdaAR & 13.24 / 10.85 & 13.03 / 20.26 & 38.46 / 20.00 & 11.70 / 15.39 & 4.71 / 5.14 \\
 & TT1D & tjAR & 13.83 / 8.93 & 12.73 / 14.11 & 40.00 / 20.00 & 10.45 / 12.25 & 2.53 / 2.39 \\
 & TT3D & AdaAR (a) & 13.50 / 19.52 & 17.65 / 16.62 & 39.45 / 65.71 & 12.85 / 30.16 & 2.26 / 10.38 \\
 & TT3D & AdaAR (b) & 13.50 / 18.33 & 17.65 / 17.42 & 39.45 / 51.43 & 12.85 / 28.67 & 2.26 / 9.26 \\
 & TT3D & tjAR & 13.10 / 19.52 & 13.92 / 17.80 & 24.07 / 57.36 & 10.42 / 27.48 & 4.16 / 9.40 \\
\bottomrule
\end{tabular}
\end{adjustbox}
\end{table}

\begin{table}[htbp]\centering\scriptsize
\caption{Decoder size and initialization for DisMo + AdaAR with dual augmentation, all six probes.}
\label{tab:app_dismo_decoder}
\begin{minipage}[t]{0.495\linewidth}\centering
\setlength{\tabcolsep}{3pt}\renewcommand{\arraystretch}{0.88}
\begin{adjustbox}{max width=\linewidth}
\begin{tabular}{llrrrrr}
\toprule
Probe & Decoder & HMDB51 & ARID & IARD & Jester & SSv2 \\
\midrule
$k$NN & S imgnet & 16.74 & \underline{12.56} & \textbf{68.68} & \underline{12.27} & \textbf{2.91} \\
 & S video & \textbf{17.47} & \textbf{12.67} & 52.31 & 12.16 & \underline{2.78} \\
 & B imgnet & 15.09 & 10.15 & \underline{52.53} & \textbf{12.37} & 2.70 \\
 & B video & \underline{17.01} & 7.27 & 43.85 & 11.27 & 2.47 \\
\midrule
linear & S imgnet & \underline{23.69} & \underline{16.57} & \underline{75.82} & \underline{20.74} & \underline{6.59} \\
 & S video & \textbf{24.95} & \textbf{18.23} & 67.58 & 19.65 & 6.09 \\
 & B imgnet & 20.38 & 14.86 & \textbf{79.45} & \textbf{22.63} & \textbf{7.05} \\
 & B video & 20.12 & 14.11 & 50.66 & 20.21 & 5.92 \\
\midrule
linear$_{\mathrm{lw}}$ & S imgnet & \underline{23.69} & \underline{16.52} & \underline{76.70} & \underline{22.58} & \underline{6.79} \\
 & S video & \textbf{25.15} & \textbf{18.07} & 67.80 & 20.97 & 6.27 \\
 & B imgnet & 20.85 & 14.86 & \textbf{80.33} & \textbf{23.22} & \textbf{7.20} \\
 & B video & 20.25 & 14.11 & 51.76 & 21.87 & 6.03 \\
\bottomrule
\end{tabular}
\end{adjustbox}
\end{minipage}\hfill
\begin{minipage}[t]{0.495\linewidth}\centering
\setlength{\tabcolsep}{3pt}\renewcommand{\arraystretch}{0.88}
\begin{adjustbox}{max width=\linewidth}
\begin{tabular}{llrrrrr}
\toprule
Probe & Decoder & HMDB51 & ARID & IARD & Jester & SSv2 \\
\midrule
MLP & S imgnet & 22.57 & \underline{19.99} & \textbf{87.03} & \underline{34.17} & 9.91 \\
 & S video & \textbf{26.21} & \textbf{20.52} & 70.99 & 30.90 & \underline{10.05} \\
 & B imgnet & \underline{25.22} & 19.78 & \underline{83.52} & \textbf{34.98} & \textbf{10.89} \\
 & B video & 22.30 & \underline{19.99} & 67.69 & 33.40 & 9.90 \\
\midrule
MLP$_{\mathrm{lw}}$ & S imgnet & 23.83 & \textbf{22.61} & \textbf{87.47} & \underline{35.31} & \underline{10.73} \\
 & S video & \textbf{26.74} & \underline{21.17} & 71.54 & 33.31 & 10.56 \\
 & B imgnet & \underline{25.08} & 19.29 & \underline{83.74} & \textbf{35.96} & \textbf{11.21} \\
 & B video & 22.30 & 18.65 & 67.25 & 34.93 & 10.21 \\
\midrule
attentive & S imgnet & 22.57 & \textbf{22.45} & \textbf{89.89} & 46.95 & \underline{13.33} \\
 & S video & \textbf{25.35} & 19.35 & 79.01 & \underline{47.31} & 13.23 \\
 & B imgnet & 23.03 & \underline{20.84} & \underline{85.60} & \textbf{47.90} & \textbf{13.91} \\
 & B video & \underline{23.49} & 18.17 & 76.92 & 46.28 & 12.52 \\
\bottomrule
\end{tabular}
\end{adjustbox}
\end{minipage}
\end{table}

\begin{table}[htbp]\centering\scriptsize
\caption{TT3D + Diff Compression with Kinetics-710 added to the pretraining mixture (6 epochs on the larger mixture against 8 epochs on OpenVid and Moments-in-Time).}
\label{tab:app_k710}
\begin{minipage}[t]{0.495\linewidth}\centering
\setlength{\tabcolsep}{3pt}\renewcommand{\arraystretch}{0.88}
\begin{adjustbox}{max width=\linewidth}
\begin{tabular}{lllrrrrr}
\toprule
Probe & Decoder & Data & HMDB51 & ARID & IARD & Jester & SSv2 \\
\midrule
$k$NN & S imgnet & OpenVid+MiT & 9.79 & 16.21 & \textbf{76.70} & 18.21 & 4.63 \\
 & S imgnet & +K710 & 7.02 & 13.08 & 73.08 & 12.55 & 2.71 \\
 & S video & OpenVid+MiT & \textbf{13.57} & \textbf{21.81} & 62.20 & \textbf{27.05} & \textbf{9.11} \\
 & S video & +K710 & \underline{11.25} & \underline{17.11} & \underline{73.74} & \underline{21.33} & \underline{7.49} \\
\bottomrule
\end{tabular}
\end{adjustbox}
\end{minipage}\hfill
\begin{minipage}[t]{0.495\linewidth}\centering
\setlength{\tabcolsep}{3pt}\renewcommand{\arraystretch}{0.88}
\begin{adjustbox}{max width=\linewidth}
\begin{tabular}{lllrrrrr}
\toprule
Probe & Decoder & Data & HMDB51 & ARID & IARD & Jester & SSv2 \\
\midrule
attentive & S imgnet & OpenVid+MiT & 12.64 & 21.98 & 67.03 & \underline{53.89} & 18.28 \\
 & S imgnet & +K710 & 7.28 & 13.67 & \textbf{82.64} & 31.28 & 11.88 \\
 & S video & OpenVid+MiT & \textbf{25.15} & \textbf{37.47} & 74.84 & \textbf{73.25} & \textbf{25.92} \\
 & S video & +K710 & \underline{15.88} & \underline{28.79} & \underline{79.89} & 44.13 & \underline{18.43} \\
\bottomrule
\end{tabular}
\end{adjustbox}
\end{minipage}
\end{table}

\subsection{Final Comparison and Every Run}
\label{sec:appendix-complete-final}

Table~\ref{tab:app_final_probes} extends Table~\ref{tab:final} to every probe and to V-JEPA~2 at two frame rates. Table~\ref{tab:app_all_runs} lists every pretrained run of this work on every task.

\begin{table}[htbp]\centering\scriptsize
\caption{Final comparison under all six frozen probes. V-JEPA~2 is evaluated with 8 frames at 6 fps and 16 frames at 12 fps. The single-frame DINOv3 row is the appearance reference (attentive probe only). Best and second-best per probe and column marked, the reference excluded.}
\label{tab:app_final_probes}
\begin{minipage}[t]{0.495\linewidth}\centering
\setlength{\tabcolsep}{3pt}\renewcommand{\arraystretch}{0.88}
\begin{adjustbox}{max width=\linewidth}
\begin{tabular}{llrrrrr}
\toprule
Probe & Method & HMDB51 & ARID & IARD & Jester & SSv2 \\
\midrule
$k$NN & V-JEPA~2 (8f) & \underline{19.26} & 7.91 & 61.76 & 11.24 & 3.94 \\
 & V-JEPA~2 (16f) & 18.07 & 10.04 & \textbf{73.52} & 9.76 & 3.85 \\
 & VideoMAE & \textbf{20.19} & \underline{17.50} & 63.96 & \underline{14.30} & \underline{4.83} \\
 & DisMo (dual aug) & 16.74 & 12.56 & \underline{68.68} & 12.27 & 2.91 \\
 & \textbf{TT-VidT} & 13.57 & \textbf{21.81} & 62.20 & \textbf{27.05} & \textbf{9.11} \\
\midrule
linear & V-JEPA~2 (8f) & 25.88 & 16.51 & 61.98 & 21.98 & 9.33 \\
 & V-JEPA~2 (16f) & \underline{25.94} & 17.01 & \underline{74.73} & \underline{22.61} & 9.82 \\
 & VideoMAE & \textbf{29.65} & \underline{21.98} & 72.31 & \underline{22.61} & \underline{9.94} \\
 & DisMo (dual aug) & 23.69 & 16.57 & \textbf{75.82} & 20.74 & 6.59 \\
 & \textbf{TT-VidT} & 23.23 & \textbf{27.69} & 64.73 & \textbf{38.82} & \textbf{14.44} \\
\midrule
linear$_{\mathrm{lw}}$ & V-JEPA~2 (8f) & 24.88 & 15.66 & 70.55 & 23.00 & 9.69 \\
 & V-JEPA~2 (16f) & \underline{25.74} & 18.50 & \underline{76.37} & \underline{23.20} & 9.89 \\
 & VideoMAE & \textbf{29.12} & \underline{22.48} & 71.10 & 22.67 & \underline{10.41} \\
 & DisMo (dual aug) & 23.69 & 16.52 & \textbf{76.70} & 22.58 & 6.79 \\
 & \textbf{TT-VidT} & 23.10 & \textbf{27.79} & 65.38 & \textbf{42.73} & \textbf{15.22} \\
\bottomrule
\end{tabular}
\end{adjustbox}
\end{minipage}\hfill
\begin{minipage}[t]{0.495\linewidth}\centering
\setlength{\tabcolsep}{3pt}\renewcommand{\arraystretch}{0.88}
\begin{adjustbox}{max width=\linewidth}
\begin{tabular}{llrrrrr}
\toprule
Probe & Method & HMDB51 & ARID & IARD & Jester & SSv2 \\
\midrule
MLP & V-JEPA~2 (8f) & \underline{23.83} & 13.97 & 68.90 & 25.99 & 9.09 \\
 & V-JEPA~2 (16f) & 22.70 & 14.77 & \underline{83.08} & 26.42 & 10.04 \\
 & VideoMAE & \textbf{27.93} & \underline{22.13} & 74.73 & 31.94 & \underline{12.59} \\
 & DisMo (dual aug) & 22.57 & 19.99 & \textbf{87.03} & \underline{34.17} & 9.91 \\
 & \textbf{TT-VidT} & 21.24 & \textbf{30.89} & 67.80 & \textbf{49.39} & \textbf{18.68} \\
\midrule
MLP$_{\mathrm{lw}}$ & V-JEPA~2 (8f) & 22.57 & 14.07 & 63.85 & 25.50 & 9.22 \\
 & V-JEPA~2 (16f) & 22.10 & 14.87 & \underline{81.87} & 26.59 & 10.22 \\
 & VideoMAE & \textbf{26.74} & 21.63 & 76.48 & 32.04 & \underline{12.54} \\
 & DisMo (dual aug) & \underline{23.83} & \underline{22.61} & \textbf{87.47} & \underline{35.31} & 10.73 \\
 & \textbf{TT-VidT} & 20.98 & \textbf{29.45} & 67.58 & \textbf{53.39} & \textbf{19.33} \\
\midrule
attentive & V-JEPA~2 (8f) & 21.38 & 13.67 & 68.35 & 27.31 & 9.71 \\
 & V-JEPA~2 (16f) & 22.70 & 16.01 & 79.34 & 27.41 & 10.59 \\
 & VideoMAE & \textbf{27.73} & \underline{24.37} & \underline{80.44} & 39.47 & \underline{14.95} \\
 & DisMo (dual aug) & 22.57 & 22.45 & \textbf{89.89} & \underline{46.95} & 13.33 \\
 & \textbf{TT-VidT} & \underline{25.15} & \textbf{37.47} & 74.84 & \textbf{73.25} & \textbf{25.92} \\
 & DINOv3 1 frame & 57.51 & 22.92 & 59.56 & 62.37 & - \\
\bottomrule
\end{tabular}
\end{adjustbox}
\end{minipage}
\end{table}

\begin{table}[htbp]\centering\scriptsize
\caption{Every pretrained run of this work, frozen attentive probe on every task and end-to-end finetune (FT) where run. Best and second-best per column marked. Decoder is size S pretrained on ImageNet unless named. TT3D + AdaAR with dual augmentation was trained twice, with (b) and without (a) a final LayerNorm in the decoder.}
\label{tab:app_all_runs}
\setlength{\tabcolsep}{2.5pt}\renewcommand{\arraystretch}{0.88}
\begin{adjustbox}{max width=\linewidth}
\begin{tabular}{lrrrrrrrrrr}
\toprule
Run & HMDB51 & ARID & IARD & Jester & SSv2 & D48 & D48 FT & EK-V & EK-V FT & Antic. \\
\midrule
ViT3D, MAE & \underline{27.73} & 24.37 & 80.44 & 39.47 & 14.95 & 8.68 & 8.43 & \textbf{35.62} & - & 22.81 \\
ViT3D, AdaAR & 16.68 & 13.58 & 53.41 & 12.85 & 3.99 & 7.11 & 6.29 & 26.43 & - & 23.47 \\
ViT3D, AdaAR, dual aug & 22.96 & 17.64 & 70.77 & 32.62 & 9.78 & 9.34 & 7.77 & 28.62 & - & 23.33 \\
ViT3D, tjAR & 19.32 & 15.66 & 62.86 & 13.01 & 4.86 & 7.87 & 6.45 & 27.58 & - & 22.97 \\
ViT3D, tjAR, dual aug & 20.78 & 17.85 & 72.86 & 30.61 & 10.05 & 8.43 & 8.73 & 29.78 & - & 23.59 \\
ViT3D, AR & 18.46 & 14.12 & 54.95 & 12.85 & 3.93 & 8.83 & 7.66 & 25.48 & - & 22.62 \\
ViT3D, MAE-Diff & \textbf{27.93} & 18.75 & 74.95 & 31.68 & 11.73 & 8.93 & 7.16 & \underline{33.62} & - & 23.53 \\
ViT3D, DiffComp & 20.19 & 17.70 & 77.03 & 19.82 & 6.57 & 7.16 & 5.84 & 29.24 & - & 23.39 \\
DisMo, MAE & 17.54 & 18.15 & 73.19 & 29.71 & 11.20 & 8.17 & 5.28 & 27.77 & - & 22.51 \\
DisMo, AdaAR & 17.87 & 15.46 & 56.81 & 16.46 & 5.29 & 9.19 & 6.29 & 25.20 & - & 23.30 \\
DisMo, AdaAR, dual aug & 22.57 & 22.45 & \textbf{89.89} & 46.95 & 13.33 & 9.19 & 8.12 & 31.97 & - & 22.72 \\
DisMo, AdaAR, B imgnet, dual aug & 23.03 & 20.84 & \underline{85.60} & 47.90 & 13.91 & 9.44 & 7.11 & 32.75 & - & \underline{23.70} \\
DisMo, AdaAR, B video, dual aug & 23.49 & 18.17 & 76.92 & 46.28 & 12.52 & 7.51 & 6.60 & 30.93 & - & 22.60 \\
DisMo, AdaAR, S video, dual aug & 25.35 & 19.35 & 79.01 & 47.31 & 13.23 & \textbf{10.15} & 7.72 & 31.69 & - & \textbf{23.77} \\
DisMo, tjAR & 17.74 & 14.87 & 58.79 & 15.94 & 4.35 & 8.17 & 6.45 & 25.48 & - & 22.86 \\
DisMo, tjAR, dual aug & 24.49 & 21.49 & \underline{85.60} & 48.23 & 13.10 & 8.48 & 8.88 & 32.00 & - & 23.62 \\
DisMo, AR & 16.48 & 14.02 & 57.25 & 13.54 & 3.54 & 8.58 & 6.70 & 23.41 & - & 22.81 \\
DisMo, MAE-Diff & 18.86 & 15.71 & 54.84 & 21.66 & 7.89 & 8.32 & 5.43 & 29.79 & - & 21.99 \\
DisMo, DiffComp & 15.42 & 16.51 & 59.12 & 12.98 & 5.53 & 6.70 & 6.65 & 25.33 & - & 22.47 \\
TT1D, MAE & 10.06 & 17.65 & 62.42 & 26.65 & 10.05 & 5.69 & 10.91 & 24.28 & 37.92 & - \\
TT1D, AdaAR & 13.24 & 13.03 & 38.46 & 11.70 & 4.71 & 7.06 & 9.04 & 22.79 & 37.82 & - \\
TT1D, AdaAR, dual aug & 10.85 & 20.26 & 20.00 & 15.39 & 5.14 & 7.16 & 8.22 & 21.69 & 33.97 & - \\
TT1D, tjAR & 13.83 & 12.73 & 40.00 & 10.45 & 2.53 & 7.06 & 8.58 & 21.34 & 36.07 & - \\
TT1D, tjAR, dual aug & 8.93 & 14.11 & 20.00 & 12.25 & 2.39 & 5.94 & 5.69 & 19.64 & 26.14 & - \\
TT1D, AR & 13.24 & 13.38 & 28.35 & 11.86 & 4.86 & 6.14 & 7.66 & 22.87 & 38.54 & - \\
TT1D, MAE-Diff & 6.88 & 14.52 & 57.14 & 15.68 & 6.00 & 5.18 & 13.71 & 24.46 & 39.24 & - \\
TT1D, DiffComp & 8.74 & 16.41 & 76.48 & 24.32 & 7.91 & 5.58 & 10.61 & 22.75 & 38.23 & - \\
TT3D, MAE & 7.61 & 11.49 & 24.07 & 11.07 & 4.98 & 6.90 & 15.38 & 22.72 & 42.78 & 20.30 \\
TT3D, AdaAR & 13.50 & 17.65 & 39.45 & 12.85 & 2.26 & 7.26 & 9.04 & 21.85 & 36.61 & 20.85 \\
TT3D, AdaAR, S rand+diff & 17.41 & 12.33 & 34.07 & 9.76 & 3.29 & 6.50 & 14.62 & 21.72 & 40.84 & 20.62 \\
TT3D, AdaAR, S rand+reg & 18.20 & 10.69 & 57.47 & 11.07 & 5.63 & 7.97 & 13.20 & 24.14 & 38.92 & 21.35 \\
TT3D, AdaAR, dual aug, (a) & 19.52 & 16.62 & 65.71 & 30.16 & 10.38 & 7.87 & 15.28 & 26.14 & \textbf{45.38} & 22.08 \\
TT3D, AdaAR, dual aug, (b) & 18.33 & 17.42 & 51.43 & 28.67 & 9.26 & 8.07 & 16.45 & 24.28 & 43.42 & 22.01 \\
TT3D, tjAR & 13.10 & 13.92 & 24.07 & 10.42 & 4.16 & 8.88 & 11.42 & 21.28 & 37.29 & 20.94 \\
TT3D, tjAR, dual aug & 19.52 & 17.80 & 57.36 & 27.48 & 9.40 & 7.31 & 16.24 & 25.04 & 43.61 & 22.20 \\
TT3D, AR & 12.84 & 15.76 & 23.96 & 11.83 & 4.39 & 6.29 & 10.30 & 22.07 & 36.19 & 21.12 \\
TT3D, MAE-Diff & 6.68 & 7.71 & 32.09 & 10.32 & 4.82 & 6.45 & 15.99 & 24.19 & \underline{43.80} & 21.13 \\
TT3D, DiffComp & 12.64 & 21.98 & 67.03 & 53.89 & 18.28 & 5.63 & 16.35 & 28.23 & 42.49 & 19.34 \\
TT3D, DiffComp, +K710 & 7.28 & 13.67 & 82.64 & 31.28 & 11.88 & 5.74 & 15.08 & 26.87 & 41.19 & 19.53 \\
TT3D, DiffComp, B imgnet & 9.20 & 19.99 & 83.19 & 61.00 & 17.00 & 6.09 & 16.04 & 26.56 & 41.76 & 20.04 \\
TT3D, DiffComp, B video & 23.16 & \textbf{37.84} & 77.36 & \underline{69.99} & \underline{23.99} & \underline{9.64} & 17.41 & 32.82 & 42.31 & 21.62 \\
TT3D, DiffComp, L imgnet & 6.15 & 14.52 & 76.37 & 26.95 & 8.84 & 5.69 & 13.60 & 24.99 & 40.60 & 20.47 \\
TT3D, DiffComp, L video & 11.98 & 17.90 & 74.73 & 34.37 & 12.15 & 7.16 & \underline{17.51} & 27.52 & 39.42 & 20.65 \\
TT3D, DiffComp, S rand+diff & 7.61 & 13.18 & 79.01 & 28.46 & 11.71 & 6.70 & 16.60 & 23.88 & 43.18 & 19.48 \\
TT3D, DiffComp, S rand+reg & 4.57 & 13.28 & 71.76 & 20.21 & 7.53 & 5.53 & 15.13 & 24.89 & 41.68 & 20.50 \\
TT3D, DiffComp, S video & 25.15 & \underline{37.47} & 74.84 & \textbf{73.25} & \textbf{25.92} & 8.17 & \textbf{18.63} & 32.54 & 40.68 & 21.33 \\
TT3D, DiffComp, S video, +K710 & 15.88 & 28.79 & 79.89 & 44.13 & 18.43 & 8.53 & \underline{17.51} & 32.76 & 41.44 & 20.82 \\
\bottomrule
\end{tabular}
\end{adjustbox}
\end{table}

\section{Motion Probe, Attribution Controls and Robustness}
\label{sec:appendix-review}

\begin{figure}[htbp]\centering
\includegraphics[width=\textwidth]{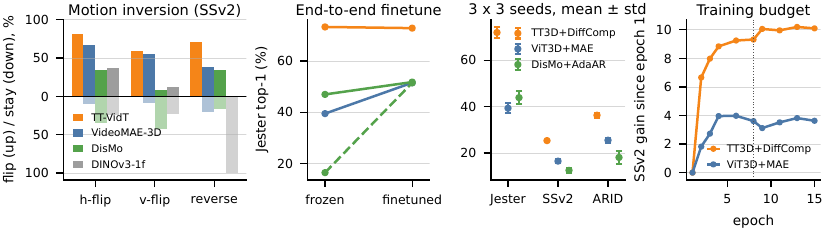}
\caption{From left: the motion-inversion probe on SSv2 (flip above the axis, stay below), frozen probe against end-to-end finetuning on Jester (dashed: DisMo without augmentation), the canonical cells over three pretraining and three probe seeds, and the cumulative SSv2 gain over a 15-epoch continuation (the dotted line marks the 8-epoch budget).}
\label{fig:app_review_panels}
\end{figure}

This section collects the experiments added after submission. Figure~\ref{fig:app_review_panels} summarizes them.

\subsection{Motion-Inversion Probe}
\label{sec:appendix-review-flip}

SSv2 contains class pairs that are exact mirrors under a pixel transform (Table~\ref{tab:app_rb_pairs}): three pairs under horizontal flip (left and right), two under vertical flip (up and down), and four under time reversal (towards and away, closer and apart). Time reversal changes only the frame order. For each frozen encoder we train an attentive probe on clean features without flip or reversal augmentation, keep the validation clips the probe classifies correctly, transform the raw frames, re-encode them, and classify with the same probe. A flip is a move to the mirror class and shows the encoder re-read the motion. A stay keeps the original class and shows the decision came from appearance, since the content is otherwise identical. The single-frame DINOv3 probe is the appearance-only control. Only pretrained checkpoints are probed, since finetuning would mix in supervision. The TT-VidT and VideoMAE rows use matched probe data of 300 clips per class, and a second TT-VidT pretraining run reproduces the pattern.

\begin{table}[htbp]\centering\scriptsize
\caption{The nine SSv2 class pairs of the motion-inversion probe (18 classes, SSv2 label ids in parentheses). The transform maps a clip of one class onto the motion of the other.}
\label{tab:app_rb_pairs}
\setlength{\tabcolsep}{3pt}\renewcommand{\arraystretch}{0.88}
\begin{adjustbox}{max width=\linewidth}
\begin{tabular}{lll}
\toprule
Transform & Class A & Class B \\
\midrule
Horizontal flip & Pushing something from left to right (93) & Pushing something from right to left (94) \\
 & Pulling something from left to right (86) & Pulling something from right to left (87) \\
 & Turning the camera left while filming something (166) & Turning the camera right while filming something (167) \\
\midrule
Vertical flip & Moving something up (45) & Moving something down (43) \\
 & Turning the camera upwards while filming something (168) & Turning the camera downwards while filming something (165) \\
\midrule
Time reversal & Moving something towards the camera (44) & Moving something away from the camera (41) \\
 & Moving something and something closer to each other (37) & Moving something and something away from each other (36) \\
 & Moving something closer to something (42) & Moving something away from something (40) \\
 & Approaching something with your camera (0) & Moving away from something with your camera (32) \\
\bottomrule
\end{tabular}
\end{adjustbox}
\end{table}

\begin{table}[htbp]\scriptsize
\begin{minipage}[t]{0.52\linewidth}\centering
\caption{Motion-inversion probe on SSv2 validation clips, flip / stay (\%). A flip moves the prediction to the mirror class after the input transform, a stay keeps the original class. Shuffle drop is the relative accuracy drop of a probe retrained on temporally shuffled features. DINOv3-1f is the single-frame appearance control.}
\label{tab:app_rb_flip}
\setlength{\tabcolsep}{3pt}\renewcommand{\arraystretch}{0.88}
\begin{adjustbox}{max width=\linewidth}
\begin{tabular}{lcccr}
\toprule
Model & Horizontal flip & Vertical flip & Time reversal & Shuffle drop \\
\midrule
\textbf{TT-VidT} & \textbf{81.2 / 0.6} & \textbf{59.5 / 0.9} & \textbf{71.4 / 1.0} & 50.9\% \\
VideoMAE-3D & 66.9 / 9.7 & 55.1 / 9.1 & 38.6 / 20.9 & 50.0\% \\
DisMo & 34.5 / 35.0 & 8.4 / 43.2 & 34.5 / 16.4 & 31.9\% \\
DINOv3-1f & 37.1 / 24.5 & 11.8 / 22.7 & 0.0 / 100.0 & 3.1\% \\
\bottomrule
\end{tabular}
\end{adjustbox}
\end{minipage}\hfill
\begin{minipage}[t]{0.455\linewidth}\centering
\caption{Frozen attentive probe against 30 epochs of end-to-end finetuning on Jester and SSv2, one shared recipe with identical encoder and head learning rates.}
\label{tab:app_rb_e2e}
\setlength{\tabcolsep}{3pt}\renewcommand{\arraystretch}{0.88}
\begin{adjustbox}{max width=\linewidth}
\begin{tabular}{lrrrr}
\toprule
 & \multicolumn{2}{c}{Jester} & \multicolumn{2}{c}{SSv2} \\
\cmidrule(lr){2-3}\cmidrule(lr){4-5}
Model & frozen & FT & frozen & FT \\
\midrule
ViT3D+MAE & 39.47 & 51.50 & 14.95 & 13.40 \\
DisMo+AdaAR (no aug) & 16.46 & 51.46 & 5.29 & 12.31 \\
DisMo+AdaAR (dual aug) & 46.95 & 51.76 & 13.33 & 14.85 \\
TT-VidT & 73.25 & 72.79 & 25.92 & 27.72 \\
\bottomrule
\end{tabular}
\end{adjustbox}
\end{minipage}
\end{table}

TT-VidT keeps the original class on at most 1\% of clips on every axis, while every baseline keeps it on 9\% to 43\%, most under time reversal. The control scores 0.0 flip and 100.0 stay under reversal, as a representation without access to frame order must. The shuffle column shows that TT-VidT and VideoMAE both use frame order, and the flip test shows that only TT-VidT reads its direction faithfully.

\subsection{Attribution Controls}
\label{sec:appendix-review-controls}

Table~\ref{tab:app_rb_controls} isolates the two ingredients TT-VidT shares with other work. A 3D ViT initialized from DINOv3 lands on the scratch ViT3D + MAE row for every motion-heavy dataset and moves only IARD, so the DINOv3 initialization does not create the motion regime. The initialization covers 12 of the 24 layers, the depth of the DINOv3 ViT-B teacher. A VTok-style motion token computed as an explicit feature difference \citep{wang2026vtokunifiedvideotokenizer} already beats every non-TT configuration on Jester, which credits the keyframe factorization, and the learned token of TT-VidT improves on it on all five datasets.

\begin{table}[htbp]\scriptsize
\begin{minipage}[t]{0.5\linewidth}\centering
\caption{Attribution controls, frozen attentive probe. Top: initialization control, a 24-layer 3D ViT whose first 12 layers are initialized from DINOv3 ViT-B (the depth of the teacher) with 3D RoPE, trained exactly as ViT3D + MAE. Bottom: a VTok-style explicit feature-difference motion token against the learned token under the same encoder, decoder and recipe.}
\label{tab:app_rb_controls}
\setlength{\tabcolsep}{3pt}\renewcommand{\arraystretch}{0.88}
\begin{adjustbox}{max width=\linewidth}
\begin{tabular}{lrrrrr}
\toprule
Model & HMDB51 & ARID & IARD & Jester & SSv2 \\
\midrule
ViT3D+MAE (scratch) & 27.73 & 24.37 & 80.44 & 39.47 & 14.95 \\
ViT3D+MAE (DINOv3 init) & 21.58 & 24.86 & 86.70 & 39.53 & 14.00 \\
\midrule
TT-VidT & 25.15 & 37.47 & 74.84 & 73.25 & 25.92 \\
\midrule
TT3D, explicit feature diff & 20.3 & 26.5 & 50.5 & 64.7 & 24.8 \\
TT3D, learned token (TT-VidT) & 25.15 & 37.47 & 74.84 & 73.25 & 25.92 \\
\bottomrule
\end{tabular}
\end{adjustbox}
\end{minipage}\hfill
\begin{minipage}[t]{0.475\linewidth}\centering
\caption{Continuation from 8 to 15 epochs under the same recipe with a fresh warmup and cosine schedule for both cells, mean $\pm$ standard deviation over 3 probe seeds. The 8-epoch rows re-probe the paper checkpoints, so they differ slightly from the 9-measurement means of Table~\ref{tab:app_rb_seeds}.}
\label{tab:app_rb_budget}
\setlength{\tabcolsep}{3pt}\renewcommand{\arraystretch}{0.88}
\begin{adjustbox}{max width=\linewidth}
\begin{tabular}{lrrrr}
\toprule
Model & Epochs & Jester & SSv2 & ARID \\
\midrule
TT3D+DiffComp & 8 & 74.23 $\pm$ 0.59 & 25.86 $\pm$ 0.16 & 36.22 $\pm$ 0.82 \\
 & 15 & 75.20 $\pm$ 0.24 & 26.73 $\pm$ 0.14 & 36.60 $\pm$ 0.81 \\
\midrule
ViT3D+MAE & 8 & 40.52 $\pm$ 0.59 & 15.34 $\pm$ 0.23 & 24.50 $\pm$ 0.80 \\
 & 15 & 43.62 $\pm$ 0.75 & 16.99 $\pm$ 0.26 & 24.68 $\pm$ 0.48 \\
\bottomrule
\end{tabular}
\end{adjustbox}
\end{minipage}
\end{table}

\subsection{End-to-End Finetuning}
\label{sec:appendix-review-e2e}

Table~\ref{tab:app_rb_e2e} finetunes the whole encoder on Jester and SSv2. The baselines gain 12 to 35 points from supervision and converge near 51.5 on Jester, while TT-VidT stays at its frozen accuracy and leads by 21.0 points on Jester and 12.9 on SSv2.

\subsection{Seeds and Training Budget}
\label{sec:appendix-review-robust}

Table~\ref{tab:app_rb_seeds} repeats the three canonical cells with three pretraining seeds and three probe seeds each. On Jester, SSv2 and ARID the worst TT3D + Diff Compression measurement lies above the best ViT3D + MAE measurement. Probe-seed variance is about as large as pretraining-seed variance. On IARD the variance comes mostly from which of the five actors is held out, and under one fixed actor split all models land at a similar level. Tables~\ref{tab:app_rb_budget} and~\ref{tab:app_rb_interval} continue the two cells from 8 to 15 epochs. The ordering is unchanged, the Jester gap stays above 31 points, and every TT3D interval after epoch 6 changes SSv2 by less than one point. All conclusions of this paper are stated for the matched budget of Section~\ref{sec:exp-setup}.

\begin{table}[htbp]\centering\scriptsize
\caption{The three canonical cells over 3 pretraining seeds $\times$ 3 probe seeds, mean $\pm$ standard deviation over the 9 measurements, frozen attentive probe. The last row compares the worst TT3D + Diff Compression measurement with the best ViT3D + MAE measurement on the three motion-heavy benchmarks.}
\label{tab:app_rb_seeds}
\setlength{\tabcolsep}{3pt}\renewcommand{\arraystretch}{0.88}
\begin{adjustbox}{max width=\linewidth}
\begin{tabular}{lrrrrr}
\toprule
Model & Jester & SSv2 & ARID & HMDB51 & IARD \\
\midrule
TT3D+DiffComp & 71.95 $\pm$ 2.39 & 25.33 $\pm$ 0.44 & 36.26 $\pm$ 1.16 & 24.59 $\pm$ 1.14 & 87.16 $\pm$ 3.59 \\
ViT3D+MAE & 39.31 $\pm$ 2.16 & 16.48 $\pm$ 0.98 & 25.48 $\pm$ 1.11 & 27.08 $\pm$ 0.68 & 87.22 $\pm$ 6.67 \\
DisMo+AdaAR & 43.87 $\pm$ 2.71 & 12.51 $\pm$ 0.99 & 18.07 $\pm$ 2.78 & 22.04 $\pm$ 1.70 & 79.90 $\pm$ 6.52 \\
\midrule
worst TT3D / best ViT3D+MAE & 67.78 / 43.25 & 24.83 / 17.94 & 34.34 / 27.20 & - & - \\
\bottomrule
\end{tabular}
\end{adjustbox}
\end{table}

\begin{table}[htbp]\centering\scriptsize
\caption{SSv2 attentive-probe change (points) between consecutive measured epochs of the 15-epoch continuation.}
\label{tab:app_rb_interval}
\setlength{\tabcolsep}{3pt}\renewcommand{\arraystretch}{0.88}
\begin{adjustbox}{max width=\linewidth}
\begin{tabular}{lrrrrrrrrr}
\toprule
Interval & 1$\to$2 & 2$\to$3 & 3$\to$4 & 4$\to$6 & 6$\to$8 & 8$\to$9 & 9$\to$11 & 11$\to$13 & 13$\to$15 \\
\midrule
TT3D+DiffComp & +6.66 & +1.32 & +0.86 & +0.41 & +0.07 & +0.74 & $-$0.10 & +0.24 & $-$0.10 \\
ViT3D+MAE & +1.82 & +0.91 & +1.23 & +0.02 & $-$0.37 & $-$0.49 & +0.40 & +0.30 & $-$0.19 \\
\bottomrule
\end{tabular}
\end{adjustbox}
\end{table}

\FloatBarrier

\section{Broader Impact}

TT-VidT is a representation-learning study rather than a deployed video understanding system. Its positive impact is mainly methodological: a more controlled way to test whether video SSL models use temporal change can support better benchmarks, more efficient video encoders, and downstream applications where motion cues matter, such as robotics, assistive perception, sports or skill analysis, and scientific video understanding. The same improvements could also be misused in surveillance-sensitive settings, because action recognition can reveal behavioral information about people even when identity is not the target. Our experiments do not introduce a new dataset of people, do not release a high-risk deployed system, and emphasize diagnostic limitations; any deployment should still account for privacy, consent, fairness across capture conditions, and domain-specific failure modes.

\section{Limitations and Future Work Discussion}

The main limitation is scale and statistical coverage. We study one matched pretraining mixture, one roughly 170M$\sim$190M encoder scale, and mostly single-seed runs; this is sufficient for a controlled architecture-objective comparison but not for claiming that the same profile will persist at much larger data, model, or schedule scales. Future work should test whether TT-VidT's motion-prioritized behavior survives longer pretraining, stronger data mixtures, and multi-seed evaluation. Another natural direction is to combine TT3D with DisMo-style dual augmentation, since our current comparison evaluates them as separate operating points. The appearance-vs-motion diagnostic is also intentionally lightweight: it uses a single-frame DINOv3 attentive probe as an appearance reference, so alternative appearance baselines such as DINOv2 or CLIP may shift dataset positions. Finally, selective unfreezing of the spatial encoder, richer temporal readouts, and broader evaluation on untrimmed or egocentric video tasks may clarify where compact motion channels help and where appearance-rich representations remain preferable.

\section{Data, Licenses, and Release}

We use existing datasets and pretrained components rather than introducing a new data source. OpenVid-1M \citep{DBLP:conf/iclr/NanXZFYCL0T25} is distributed for research use under CC-BY-4.0 while also requiring users to respect upstream video licenses; Moments-in-Time v2 \citep{DBLP:journals/pami/MonfortVOAZRBYB20} and the downstream benchmarks are used through their original academic or public access channels and are not redistributed by this work. Our code and reproduction instructions are released at \url{https://github.com/KohakuBlueleaf/TTVidT} under the Apache License 2.0, while raw datasets should be obtained from their original providers under the corresponding terms.

\end{document}